\documentclass[journal]{IEEEtran}
\usepackage{algorithmic,amsfonts,amsmath,amssymb,booktabs,cite,color,epsfig,float,graphicx,latexsym,makecell,multirow,palatino,soul,threeparttable}
\usepackage{url}
\usepackage[linesnumbered,ruled,vlined]{algorithm2e}
\usepackage[caption=true,font=footnotesize]{subfig}

\def\MR2{\multirow{2}[2]{*}}

\definecolor{hl}{rgb}{0.75,0.75,0.75}
\sethlcolor{hl}
\newcommand{\XG}[1]{\textcolor[rgb]{0.00,0.00,1.00}{#1}}

\usepackage{epstopdf}
\usepackage{tabularx}

\usepackage{pifont}
\usepackage[colorlinks,linkcolor=black,anchorcolor=blue,citecolor=black]{hyperref}

\begin{document}
\title{Designing Novel Cognitive Diagnosis Models via Evolutionary Multi-Objective Neural Architecture Search

\thanks{Manuscript received --. This work was supported 
	in part by the National Key Research and Development Project under Grant 2018AAA0100105 and 2018AAA0100100,   
	in part by the National Natural Science Foundation of China under Grant 61822301, 61876123, 61906001, 62136008, U21A20512, and U1804262, 
	%in part by the Hong Kong Scholars Program under Grant XJ2019035, 
	in part by the Anhui Provincial Natural Science Foundation under Grant 1808085J06 and 1908085QF271, 
	in part by the Collaborative Innovation Program of Universities in Anhui Province under Grant GXXT-2020-013,
	and in part by the State Key Laboratory of Synthetical Automation for Process Industries under Grant PAL-N201805 
	{\it (Corresponding authors: Limiao Zhang and Xingyi Zhang).}}
}

\author{Shangshang Yang,
	Haiping Ma,	
	Cheng Zhen,
Ye Tian,
%Yuanchao Liu,
Limiao Zhang,\\
Yaochu Jin, \emph{Fellow, IEEE},
and
Xingyi Zhang, \emph{Senior Member, IEEE}
% <-this % stops a space
\thanks{S. Yang and X. Zhang is with the Key Laboratory of Intelligent Computing and Signal Processing of Ministry of Education,
	School of Artificial Intelligence, Anhui University, Hefei 230039, China (email: yangshang0308@gmail.com; xyzhanghust@gmail.com).}
%\thanks{Y. Liu is with the State Key Laboratory of Synthetical Automation for Process Industries, Northeastern University, Shenyang 110819, China. He is also with the College of Information Science and Engineering, Northeastern University, Shenyang 110819, China. (e-mail: Yuanchaoliu@126.com).}
\thanks{C. Zhen, Y. Tian, and H. Ma are with the Key Laboratory of Intelligent Computing and Signal Processing of Ministry of Education, Institutes of Physical Science and Information Technology, Anhui University, Hefei 230601, China (email: zhencheng0208@gmail.com;field910921@gmail.com;hpma@ahu.edu.cn).}
\thanks{L. Zhang is with Information Materials and Intelligent Sensing Laboratory of Anhui Province, Anhui University, Hefei, 230601, Anhui, China (email: zhanglm@ahu.edu.cn).}
\thanks{Y. Jin is with the Faculty of Technology, Bielefeld Unversity, Bielefeld 33619, Germany (email:yaochu.jin@uni-bielefeld.de).}}

\markboth{IEEE TRANSACTIONS ON XXXX, VOL. X, NO. X, MM YYYY}
{Yang \MakeLowercase{\textit{et al.}}: No title}

%\IEEEpubid{0000--0000/00\$00.00~\copyright~0000 IEEE}

\maketitle

\begin{abstract}
Cognitive diagnosis plays a vital role in modern intelligent education platforms to  reveal students' proficiency in knowledge concepts for subsequent adaptive tasks. However, due to the requirement of  high model interpretability, existing manually designed cognitive diagnosis models hold too simple architectures to meet the demand of current  intelligent education systems, where the bias of human design  also limits the emergence of effective cognitive diagnosis models. In this paper, we propose to automatically design novel cognitive diagnosis models by evolutionary multi-objective neural architecture search (NAS). Specifically, we observe  existing models can be represented by  a general model handling   three given types of inputs and thus first design an expressive search space for the NAS task in cognitive diagnosis. Then, we propose  multi-objective genetic programming (MOGP) to explore  the NAS task's search space by maximizing  model performance and  interpretability. In the  MOGP design,  each architecture is transformed into a tree architecture and encoded by a tree for easy optimization,  and a tailored genetic operation based on four sub-genetic operations is devised to  generate offspring effectively. Besides, an initialization strategy is also suggested to accelerate the convergence by evolving half of the population from existing models' variants. Experiments on two real-world  datasets demonstrate that the cognitive diagnosis models searched by the proposed approach exhibit significantly better performance than existing models and also hold  as good interpretability  as human-designed models.

\iffalse

To make the search algorithm effectively generate offspring, 
containing four sub-genetic operations is devised.
Besides,  we also propose an initialization strategy to make  half of the population evolve from  existing models' variants 
to accelerate the convergence.
Experiments on two real-world  datasets demonstrate that 
the cognitive diagnosis models searched by the proposed approach exhibit significantly better performance than existing models
and also hold  good interpretability  same as human-designed models.

and then propose  the model interpretability objective to 
formulate the NAS task as a multi-objective optimization problem (MOP) for
maintaining models' performance and interpretability simultaneously.
To tackle the formulated MOP well,
we  propose a multi-objective evolutionary algorithm to explore the devised large search space,
where .

Specifically,  a novel tree-based search space is first suggested to contain not only existing
models but also other  architectures that humans have never seen  as many as possible, 
where each cognitive diagnosis model can be represented by a binary tree.
Then, an effective evolutionary algorithm is developed to explore the suggested novel search space, which considers both the performance and the interpretability of models in the search.
\fi	
\end{abstract}


\section{Introduction}
\begin{figure}[t]
	\centering
	\includegraphics[width=0.8\linewidth]{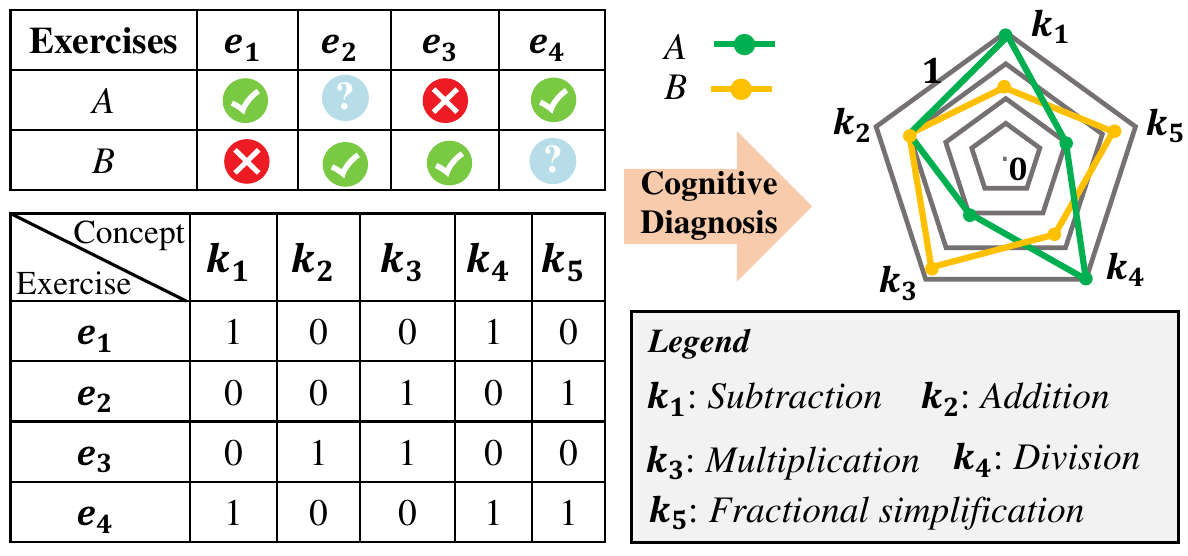}
	\caption{The illustration of cognitive diagnosis process. The response records of students $\{A,B\}$, 
		and the exercise-concept relational matrix ($Q$-matrix) are shown in the left, 
		while the top right gives  the diagnosis results.}
	\label{fig:cognitive}
\end{figure}

Cognitive diagnosis (CD) in the field of intelligent education~\cite{anderson2014engaging,burns2014intelligent} aims to
reveal students' proficiency in specific knowledge concepts
according to their historical response records of answering exercises 
and  the exercise-concept relational matrix (termed $Q$-matrix)~\cite{vanlehn1988student}. 
Fig.~\ref{fig:cognitive} gives an illustrative example of CD, 
where students $\{A,B\}$ have practiced a series of exercises (i.e., $\{e_1,e_3,e_4\}$ and $\{e_1,e_2,e_3\}$), 
and got corresponding responses. 
Based on the  records and  $Q$-matrix, 
the students' knowledge proficiency in each concept can be obtained through  CD.
By doing so, there is a wide range of intelligent education tasks, 
such as personalized exercise recommendation~\cite{Yang2023Cognitive} and targeted training~\cite{beck2007difficulties},
which can benefit from the students' diagnosis results.

% 这里还只是正常的CD分类，从传统的教育心理学 和深度学习 分类
%下一段强调这些方法的诊断函数都是手动设计的， (即使 NCD尝试去提出一个模型整合，但仍然是手动设计的)
% 手动设计NCD去整合以前的方法
%。。。。。他们不行，太耗时太耗经验去设计，因此我们想去自动设计通过NAS
% Related work 提一下现有的CD从两个方法去改进（1）改进输入 （2）改进diagnosis 函数....

With the rising demand for cognitive diagnosis models (CDMs) in  online  education platforms, 
many researchers  developed various CD approaches, which are generally grouped into two types.
The first genre of approaches  is mainly proposed by researchers in  educational psychology.
Their designed CDMs usually rely on  simple handcrafted functions to model student-exercise interactions 
and  portray the student learning ability in a one-dimensional  vector or other manners.
The representatives include  \textit{Item Response Theory} (IRT)~\cite{embretson2013item},  
\textit{Deterministic Input, Noisy ’And’ gate} (DINA)~\cite{Torre2009DINA}, \textit{Multidimensional IRT} (MIRT)~\cite{reckase2009multidimensional}, and Matrix Factorization (MF)~\cite{koren2009matrix}.
\iffalse
 \textit{Item Response Theory} (IRT)~\cite{embretson2013item} and  \textit{Deterministic Inputs, Noisy-And gate} (DINA)~\cite{Torre2009DINA} are  two pioneering approaches, 
where IRT and DINA utilize a unidimensional continuous vector and a binary vector respectively to denote the student mastery 
for predicting the probabilities of a student correctly answering exercises.
In addition, there are also some  CD approaches  improving  above two CDMs or using other techniques, 
such as MIRT~\cite{reckase2009multidimensional} which extends IRT's unidimensional student and exercise latent traits into multidimensional space, and MF~\cite{koren2009matrix} based on the matrix factorization technique.
\fi
The second genre of ones~\cite{cheng2019dirt,wang2020neural,zhou2021modeling,gao2021rcd,wang2021using} 
is  based on  neural networks (NNs), where    the student learning ability is portrayed by an inner latent vector.
The representatives contain \textit{Neural
Cognitive Diagnosis} (NCD)~\cite{wang2020neural},  
\textit{Prerequisite Attention model for Knowledge Proficiency diagnosis} (PAKP)~\cite{ma2022prerequisite}, and \textit{Relation map driven Cognitive Diagnosis}  (RCD)~\cite{gao2021rcd}
.

As the critical  components of CDMs,  
diagnostic functions are mainly responsible for  predicting student exercising scores by integrating three types of input vectors (i.e., student/exercise/concept-related input vector) in a highly interpretable manner.
To pursue  high model interpretability,
existing CDMs' diagnostic functions are desired to hold simple  architectures. 
For example, IRT~\cite{embretson2013item} and MF~\cite{koren2009matrix}  utilize the simple logistic function and   inner-product respectively as their diagnostic functions. 
However, there exist two kinds of problems for these simple handcrafted diagnostic functions.
Firstly,   simple  diagnostic functions' architectures disable CDMs from modeling  complex relationships between students and exercises well~\cite{ma2022knowledge}, failing to meet the demands of modern  education systems containing a large quantity of student exercising data.
Secondly, the design of existing diagnostic functions heavily relies on  researchers' knowledge of both educational psychology and NNs~\cite{wang2020neural}, which is  labor-intensive  and needs a lot of trial-and-error.  And the human design bias may limit the emergence of novel diagnostic functions to some extent.
Furthermore, recent CD approaches~\cite{gao2021rcd,wang2021using,ma2022prerequisite} put less focus on the architecture design of diagnostic functions  but on enhancing the input vectors  for high performance,
which hinders the development of CDMs to some extent.

For the above reasons, this paper aims to develop novel CDMs by automatically designing effective diagnostic function architectures.
Since  Zoph and Le~\cite{zoph2016neural} proposed to search  neural architectures for image tasks, 
neural architecture search (NAS)~\cite{elsken2019neural,Hao2023Relative,Sun2020Completely} has been widely applied to many research fields and  achieved significant success~\cite{liu2018darts,wang2020textnas,baruwa2019leveraging}. 
Among various search strategies of NAS, including reinforcement learning~\cite{elsken2018efficient} and gradient optimization~\cite{liu2018darts,Yuan2022Exploiting}, %and bayesian optimization~\cite{kandasamy2018neural},
evolutionary algorithms (EAs), especially  multi-objective evolutionary algorithms (MOEAs), have shown a more powerful  ability to search~\cite{elsken2019neural}. 
Moreover, compared to other NAS approaches, MOEA-based NAS approaches~\cite{liu2021survey} are superior 
in getting out of local optima and presenting trade-offs among multiple objectives%(e.g, model performance, the number of model parameters, and model inference time)
, where  many architectures holding different attributes can be found  in a single run.
The representative approaches include \textit{Neural Architecture Search using Multi-Objective
	Genetic Algorithm }(NSGA-Net)~\cite{lu2019nsga}, and \textit{Lamarckian Evolutionary algorithm for Multi-Objective Neural Architecture DEsign} (LEMONADE)~\cite{elsken2018efficient}.
%and OMOPSO~\cite{wang2019evolving
However, existing NAS approaches cannot be applied to CD due to the difference in search space between CD and other tasks, and  different search space  generally needs  different  MOEAs~\cite{liu2021survey}, 
whose  representations and genetic operations are task-tailored~\cite{tian2019evolutionary},  
 further hindering them from being  applied to CD.

%其中 基于进化的NAS方法很棒，基于MOEA的NAS特别棒，能够解决很多问题，包括。。。。 how

Therefore, this paper proposes an evolutionary multi-objective NAS to design novel CDMs (termed EMO-NAS-CD), 
where an expressive search space is first  devised  and 
 multi-objective genetic programming (MOGP) is employed to explore the search space to  develop  high-performance  CDMs with good interpretability. 
Specifically, our main contributions are as follows:
\begin{enumerate}

	\item 
	This paper is the first NAS work to  design  CDMs, which explores the search space design and  search strategy design of NAS.
	Regarding the search space, we first design an expressive  search space for the NAS task of CD (NAS-CD) by summarizing existing diagnostic function architectures. 
	Within, each candidate architecture is denoted by a general model, which takes at most three given types of input vectors as input nodes.
	Then, regarding the search strategy, we propose MOGP to explore the search space by solving a bi-objective  problem of NAS-CD, which maximizes   the objectives of  model performance  and  interpretability simultaneously.

\iffalse
To make the searched highly interpretable,
we propose to optimizethe model performance and model interpretability simultaneously and thus formulate the NAS-CD task as a bi-objective optimization problem,	where the interpretability of an architecture is intuitively characterized with its depth, breadth, and its contained computation node number.	
\fi
	
	\item 	In the MOGP design, 
we  first transform  architectures under the search space into tree architectures 
and  then encode them by trees for  easy optimization,	which can avoid the	optimization difficulties of vector-based encoding (e.g., the problem of variable-length encoding).
Based on four sub-genetic operations, a tailored genetic operation is devised for effective offspring generation in the MOGP.
Besides, 
to accelerate the MOGP's convergence, 
we further design a prior knowledge-based  initialization strategy to  evolve partial individuals of the population from existing CDMs' variants.

	\iffalse
	To avoid some optimization difficulties in general  MOEAs (e.g., variable-length encoding difficulty in \XG{vector-based encoding}),
	%To make the proposed MOEA effectively explore the devised search space, 
	each architecture is first transformed into its corresponding tree architecture and 
	then encoded by tree-based representation for easy optimization.		
	Besides, a tailored genetic operation  inspired from GP is suggested for effective offspring generation.
	On the basis of the above techniques, the  proposed MOEA turns out to be a MOGP.
	%To solve the formulated bi-objective optimization problem with individuals encoded by trees,	
	%we propose a multi-objective evolutionary algorithm (MOEA) to explore the devised large search space,
	%where.
	To accelerate the convergence of the proposed MOGP, we further design a population initialization strategy to  initialize partial individuals of the population from existing CDMs' variants.
	%to make half of the population  evolve from existing CDMs' variants instead of completely from scratch,  	
	%which can not only   but also maintain the good diversity of  population.} 
\fi

	\item To validate the effectiveness of the proposed EMO-NAS-CD, we compare it with some representative  CDMs on two popular education datasets.
	Experimental results show that EMO-NAS-CD can find a set of architectures to build CDMs, 
	which	 present trade-offs between interpretability and performance.
	 The found architectures  hold  both significantly better prediction performance and good interpretability.
	Moreover, we verify the effectiveness of the suggested genetic operation as well as the initialization strategy, 
	and we also demonstrate the superiority of the devised model interpretability objective over the common model complexity.

\end{enumerate}

The rest of this paper is as follows. 
Section II reviews  existing CD approaches and  presents the motivation for this work. 
Section III introduces the proposed search space.
Section IV presents the details of the proposed approach. 
 The experiments are shown in Section V, and we give conclusions and future work in Section VI.

\section{Preliminaries and Related Work }
%This section  presents the preliminaries and  the related work about CD  and gives the motivation of this work.

\subsection{Preliminaries of Cognitive Diagnosis Task}
Formally, there are $N$ students, $M$ exercises, and $K$ knowledge concepts in an intelligent education platform for the cognitive diagnosis task,  which can be represented by $S = \{s_1,s_2,\cdots,s_N\}$, $E= \{e_1,e_2,\cdots,e_M\}$, and $C=\{c_1,c_2,\cdots,c_K\}$, respectively. 
Besides, there is commonly an exercise-concept relation matrix $Q= (Q_{jk}\in \{0,1\})^{M\times K}$,  $Q$-matrix, to depict the relationship between  exercises and knowledge concepts, where $Q_{jk}=1$ means the exercise $e_j$ contains the knowledge concept $c_k$ and $Q_{jk}=0$ otherwise.
 $R_{log}$ is used to denote the students' exercising response logs and it can be represented by a set of triplets
$(s_i,e_j,r_{ij})$, where $s_j \in S$, $e_j \in E$, and 
$r_{ij}\in \{0,1\}$ refers to the response score of student $s_i$ on exercise $e_j$. Here $r_{ij}=1$ indicates the answer of student $s_i$ on  $e_j$ is correct and $r_{ij}=0$ otherwise.

Based on the students' response logs $R_{log}$ and $Q$-matrix, 
the  cognitive diagnosis task mines the students' proficiency in knowledge concepts  
by building a model  $\mathcal{F}$ to predict the students' exercising score.
To predict the score of student  $s_i$  on exercise $e_j$, 
the model $\mathcal{F}$ can take three types of inputs, including the student-related feature vector $\mathbf{h}_S\in R^{1\times D}$, the exercise-related feature vector $\mathbf{h}_E\in R^{1\times D}$, 
and the knowledge concept-related feature vector $\mathbf{h}_C\in R^{1\times K}$, 
which can be obtained by
\begin{equation}	
	\small
	\left.
	\begin{aligned}
		\mathbf{h}_S &= \mathbf{x}_i^S \times W_S,  W_S\in R^{N\times D}\\
		\mathbf{h}_E &= \mathbf{x}_j^E \times W_E,  W_E\in R^{M\times D}\\
		\mathbf{h}_C &= \mathbf{x}_j^E \times Q = (Q_{j1}, Q_{j2},\cdots, Q_{jK})\\
	\end{aligned}
\right.,	
\end{equation}
where $D$ is the embedding dimension (usually equal to $K$ for consistency),  
$\mathbf{x}_i^S \in \{0,1\}^{1\times N}$ is the  one-hot vector for student $s_i$,
 $\mathbf{x}_j^E \in \{0,1\}^{1\times M}$ is the  one-hot vector for exercise $e_j$,
 and $W_S$ and $W_E$ are trainable matrices in the embedding layers. 
 Then, the model $\mathcal{F}$  outputs the predicted response $ \hat{r}_{ij}$ as 
 $\hat{r}_{ij} = \mathcal{F}(\mathbf{h}_S,\mathbf{h}_E,\mathbf{h}_C)$,
where $\mathcal{F}(\cdot)$ is  the diagnostic function to combine three types of inputs in different manners. 
Generally speaking, after training the model $\mathcal{F}$ based on students' response logs, 
each bit value of  $\mathbf{h}_S$ represents the student's proficiency in the corresponding  knowledge concept.

\subsection{Related Work on Cognitive Diagnosis}\label{sec:related CD}

% 需要每一个工作  详细地用一段话介绍

In the past decades, a series of CDMs have been developed based on researchers' experiences in   educational psychology  and  deep neural networks (DNNs), mainly from two perspectives.

\subsubsection{\textbf{Incorporating  Richer Input Information}}
As introduced above, there are three types of inputs that can be used for the diagnostic function in a CDM, 
including the student-related vector $\mathbf{h}_S$, the exercise-related vector $\mathbf{h}_E$, and the knowledge concept-related vector $\mathbf{h}_C$.
Therefore, the first type of approaches aims to incorporate richer context information or other  information into these input vectors to boost the diagnostic function inputs for improving the prediction performance.

To achieve this, Zhou \emph{et al.}~\cite{zhou2021modeling} proposed \textit{Educational 	context-aware Cognitive Diagnosis} (ECD)~\cite{zhou2021modeling} to model educational context-aware features in student learning.
Specifically,  the student's educational contexts (e.g., school information, student personal interests, parents' education) are incorporated into the student-related vector $\mathbf{h}_S$ by a hierarchical attention NN. 
Then, the integrated student-related vector $\mathbf{h}_S$ will be processed by a common diagnostic function.
The incorporated educational context information can indeed improve the diagnosis performance  of different diagnostic functions, including IRT, MIRT, and NCD.

In~\cite{gao2021rcd}, Gao \emph{et al.} proposed   RCD to  incorporate the model inputs with the prior relations between knowledge concepts.
To be specific, students, exercises, and concepts are first built as a hierarchical graph.
This graph  contains a student-exercise interaction map, a concept-exercise correlation map, 
and a concept dependency map that is extracted from the prior relations between knowledge concepts.
Then, a multi-level attention NN is used to achieve node aggregation of the hierarchical graph, 
and the aggregated node features are used as three input vectors, $\mathbf{h}_S$, $\mathbf{h}_E$, and $\mathbf{h}_C$, to improve the model performance.

Similarly, Wang \emph{et al.}~\cite{wang2021using} proposed  CDGK (i.e., \textit{Cognitive DiaGnosis by Knowledge concept aggregation}) to incorporate the relations between knowledge concepts into input vectors.
Different from RCD, CDGK only builds the graph structure of knowledge concepts according to the dependency among knowledge concepts.
Only the leaf nodes in the constructed graph will be used to  aggregate the target node's features.
Finally, the aggregated knowledge concept features will be taken as the concept-related vector $\mathbf{h}_C$ used for subsequent diagnosis process.

 % DINA, IRT, MIRT, MF, and NCD
 %\XG{Big Error The dimension of IRT,  MIRT and NCD should be modified.}
 
 \subsubsection{\textbf{Designing  Diagnostic Functions}}
 The above CD approaches  only focus on incorporating extra information into input vectors, 
 and directly employ existing  diagnostic functions to handle the  enhanced input vectors for diagnosis.
 In contrast, the second type of approaches focuses on  designing powerful diagnostic functions, 
 which are responsible for  combining  input vectors in highly interpretable manners.

 As the most  typical CDM, the diagnostic function of DINA~\cite{Torre2009DINA} is to 
 first obtain two binary student and concept latent features ($\boldsymbol{\theta}, \boldsymbol{\beta}\in \{0,1\}^{1\times K}$)
 and  two exercise latent features (guessing $g\in R^{1}$ and slipping $sl\in R^{1}$) from input vectors. 
 Then, the score of student $s_i$  on exercise $e_j$ can be represented as $\hat{r}_{ij} = g^{1-nt}(1-sl)^{nt}$, 
 where $nt = \prod_{k} \boldsymbol{\theta}_{k}^{\boldsymbol{\beta}_k}$. 
 Despite the high interpretability of its diagnostic function, 
DINA suffers from poor prediction performance in  current CD tasks due to its poor scalability on large-scale student exercising data.

 % 来自NCD
  As another typical CDM, the diagnostic function of IRT~\cite{embretson2013item} first takes 
  student-related and exercise-related vectors $\mathbf{h}_S$ and $\mathbf{h}_E$, 
  and then transforms them into  one student latent feature $\theta \in R^{1}$ and 
  two exercise latent features ($\beta \in R^{1}$ and $a \in R^{1}$), respectively. 
  Next, a simple logistic function is applied to the linear transformation of $\theta$, $\beta$, and $a$, 
  e.g., a simple version is $Sigmoid(a(\theta -\beta))$ as stated in~\cite{cheng2019dirt,wang2020neural}.
  Finally,  the diagnostic function  outputs the predicted scores of the student  on  exercises.

  Similarly, MIRT~\cite{reckase2009multidimensional} applies the same  logistic function as IRT to 
  the linear transformation of  the  student latent feature $\boldsymbol{\theta} \in R^{1\times K}$, 
  the exercise latent feature ${\beta}\in R^{1}$, and the knowledge concept latent feature $\boldsymbol{\alpha}\in R^{1\times K}$.
   $\boldsymbol{\theta}$ and $\boldsymbol{\alpha}$ are equal to $\mathbf{h}_S$ and $\mathbf{h}_C$, 
  and ${\beta}$ is transformed from $\mathbf{h}_E$.
  Note that student and knowledge concept latent features in MIRT are   multidimensional
  for the demands  of multidimensional data~\cite{cheng2019dirt}.
  Finally, its prediction process can be output as $\hat{r}_{ij} = Sigmoid({\beta}+\sum \boldsymbol{\alpha}\odot \boldsymbol{\theta})$. 
Compared to IRT, MIRT exhibits better performance yet without losing interpretability.

 Differently,   MF~\cite{koren2009matrix} is originally proposed for recommender systems   but can  be used for CD from the data mining perspective, where  students and exercises in CD can correspond to users and items in recommender systems. 
 As demonstrated in~\cite{wang2020neural}, the diagnostic function of MF can be modeled as  
 directly applying the inner-product to   $\mathbf{h}_S$ and  $\mathbf{h}_E$.
 Finally, its prediction process can be represented by $\hat{r}_{ij} = \sum \mathbf{h}_S\odot \mathbf{h}_E$, 
 whose architecture is quite simple yet effective compared to other CDMs.

  The most representative  approach  NCD~\cite{wang2020neural} builds a new diagnostic function with  one shallow layer and   three fully connected (FC) layers. 
  Firstly, the student latent feature $\mathbf{f}_S\in R^{1\times K}$  and two exercise latent features
   $\mathbf{f}_{diff}\in R^{1\times K} $ and $f_{disc}\in R^{1}$ are first obtained by
   \begin{equation}
   	\small
   	\left\{
   	\begin{aligned}
   		&\mathbf{f}_S = Sigmoid(\mathbf{h}_S)\\  
   		&\mathbf{f}_{diff} = Sigmoid(\mathbf{h}_E)\\ 
   		&f_{disc} = Sigmoid(\mathbf{h}_E\times W_{disc}), W_{disc}\in R^{D\times 1}  		
   	\end{aligned} 
   \right..
   \end{equation}
Then, the shallow layer inspired by  MIRT is used to linearly combine the above features and concept-related vector $\mathbf{h}_C$ as  
$\mathbf{y} = \mathbf{h}_C\odot(\mathbf{f}_S-\mathbf{f}_{diff} )\times f_{disc}$.
Afterward, the hidden feature $\mathbf{y}$ is  fed into  three FC layers with the  monotonicity property
to get the final prediction  output. 

Ma \emph{et al.} proposed   \textit{Knowledge-Sensed Cognitive Diagnosis} (KSCD) to diagnose the student's proficiency.  
Similar to NCD,   KSCD's diagnostic function~\cite{ma2022knowledge} consists of two FC layers followed by one shallow layer. 
Two FC layers  are used to combine the learned knowledge concept features with $\mathbf{h}_S$ and $\mathbf{h}_E$, respectively, for obtaining the enhanced student  and exercise features.
Then, the shallow layer is used to further combined enhanced features and $\mathbf{h}_C$ to get the prediction.

\subsection{Motivation of This Work}

Despite the competitive performance of the above CDMs, 
their diagnosis function architectures  are too simple to model  complex student-exercise interactions well~\cite{ma2022knowledge},
especially for large-scale student exercising data in current intelligent education systems. 
Moreover,  the design of existing diagnosis function architectures  heavily relies on researcher expertise in  the domains of both education  and NNs,
which needs a lot of  trial-and-error and thus  is labor-intensive and costly~\cite{wang2020neural}.
Besides, the human design bias  may make  some potential yet beyond-human knowledge  architectures miss.
Therefore, in contrast to current CD approaches  focusing on improving model inputs, 
this paper aims to develop more effective diagnostic function architectures for CD.

\iffalse
As an automated  neural architecture design paradigm~\cite{liu2018darts,zoph2018learning},
NAS has been widely used for many research domains~\cite{elsken2019neural} and made significant progress 
since it was first proposed by Zoph and Le in~\cite{zoph2016neural}. 
Existing NAS approaches have made great achievement in various domains  to 
search the best architectures of prevailing various DNNs, including convolution neural networks (CNNs) for computer vision (CV) tasks~\cite{yu2020bignas}, 
recurrent neural networks (RNNs) for natural language process (NLP)~\cite{wang2020textnas} and speech-related~\cite{baruwa2019leveraging} tasks, 
graph neural networks (GNNs) for non-European data tasks~\cite{gao2021graph,oloulade2021graph}, 
and Transformers for  CV~\cite{su2021vision,chen2021glit},  NLP~\cite{wang2020hat,xu2021bert}, and speech-related~\cite{shi2021darts,luo2021lightspeech} tasks.
\fi

As an automated  neural architecture design paradigm~\cite{liu2018darts},
NAS has been widely used for many research domains~\cite{elsken2019neural} and made significant progress~\cite{zoph2016neural}. 
Existing NAS approaches have been used  to search the best architectures of prevailing various DNNs, including convolution neural networks (CNNs) for computer vision (CV) tasks~\cite{Li2021Automatic,yu2020bignas,Zhou2022Attention}, 
recurrent neural networks (RNNs) for natural language processing (NLP)~\cite{wang2020textnas} and speech-related~\cite{baruwa2019leveraging} tasks,
graph neural networks (GNNs) for the tasks having non-Euclidean data~\cite{oloulade2021graph}, 
and Transformers for  CV~\cite{su2021vision},  NLP~\cite{feng2021evolving}, and speech-related~\cite{luo2021lightspeech} tasks.

However,  due to the difference in search space among different domains, 
these NAS approaches cannot be  applied to search the optimal diagnostic function architecture. 
Besides,   the architectures of existing diagnostic functions can be seen as a general model,
which is used to handle   three given types of inputs and output a scalar or a vector.
To this end, this paper an evolutionary multi-objective optimization-based NAS approach for 
automatically designing effective diagnostic function architectures to build novel CDMs.
Here, we first design  an expressive  search space by summarizing existing architectures,
and then we propose MOGP to explore the devised search space by  optimizing the  objectives of model performance and model interpretability simultaneously.
To the best of our knowledge, our work is the first to apply the NAS technique to the CD task.

\begin{table}[t]
		\renewcommand{\arraystretch}{0.1}
		\centering
	\caption{{15 Candidate operators.}}

	\setlength{\tabcolsep}{0.4mm}{
		\begin{tabular}{cccc}

			%\toprule
			\toprule
			\textbf{Notation} & \textbf{Meaning} & \textbf{Syntax} & \textbf{Output shape} \\
			%\midrule
			\midrule
		%	\multicolumn{4}{c}{\textbf{Unary operators}} \\
		%	\midrule
			\textit{Neg}   & Negative & $-x,\ x\in R^{1\times D}$     & same \\
			\textit{Abs} & Absolute value & $|x|$& same\\
			\textit{Inv}&      Inverse & $1/(x+\epsilon),\ \epsilon=10^{-6}$      &same \\
			\textit{Square}&    Square   &  $x^2$     &  same\\
			\textit{Sqrt}&    Square root   &  $\mathrm{sign}(x)\cdot\sqrt{|x|+\epsilon}$     & same \\
			\textit{Tanh}&  Tanh function     &   $tanh(x)$    &  same\\
			\textit{Sigmoid}& Sigmoid function      &   $sigmoid(x)$    &  same\\
			\textit{Softplus}& Softplus function      &   $softplus(x)$    &  same\\
	
			\textit{Sum}&   Sum  of a vector     &  $\sum_i^D \mathbf{x}_i$     & single ($R^1$) \\
			\textit{Mean}&    Mean of a vector    & $(\sum_i^D \mathbf{x}_i)/ D$      & single ($R^1$)\\
			\multirow{2}[2]{*}{\textit{FFN}} &     An FC layer mapping    & $\mathbf{x}_i\times W_F,$    & \multirow{2}[2]{*}{ single ($R^1$)} \\
			&  a vector to a scalar & $W_F\in R^{D\times 1}$  & \\
			\multirow{2}[2]{*}{\textit{FFN\_D}} &     An FC layer mapping    & $\mathbf{x}_i\times W_F,$    & \multirow{2}[2]{*}{ constant ($R^D$)} \\
			&  a vector to a vector & $W_F\in R^{D\times D}$  & \\
			%\midrule
			\midrule
		%	\multicolumn{4}{c}{\textbf{Binary operators}} \\
		%	\midrule
			\textit{Add}&  Addition     &    $x+y$   &  maximum\\
			\textit{Mul}& Multiplication      & $x\cdot y$ or $x\odot y$     &  maximum\\
	
			\multirow{2}[2]{*}{\textit{Concat}}  & Concatenate  and map    & 
			$[\mathbf{x}, \mathbf{y}]\times W_{Con}$     &  \multirow{2}[2]{*}{constant ($R^{1\times D}$)}\\
			&two vectors 	& $W_{Con}\in R^{2*D\times D}$& \\
			\bottomrule
			%\bottomrule
		\end{tabular}% 
	}
	\begin{tablenotes}
		\footnotesize
		\item 'same' denotes the output shape is same as input $x$; 
		\item 'single' and 'constant' denote the output shapes are $R^1$ and $R^{1\times D}$; 
	%	\item The  denotes the output shape is $R^{1\times D}$;  	
		\item 'maximum' denotes the maximal shape between $x$ and $y$;
	%	\item The dimension $D$ is set to be equal to $K$.
		%\item The $\epsilon=10^{-6}$ is a small positive number.  
	\end{tablenotes}
	\label{tab:operator}%
\end{table}%

\section{The Proposed Search Space for CD}

%In this section, the   search space proposed for CD is first elaborated,
%and then a multi-objective optimization problem (MOP) is formulated for the NSA-CD task.

%\subsection{Search Space}
As stated above, the search space of existing NAS approaches~\cite{elsken2019neural,liu2021survey,oloulade2021graph} is task-specific, which cannot be  applied to CD for searching  diagnostic functions.

%Generally,  an effective search space commonly not only contains as many  existing  architectures as possible   but also covers a large quantity of beyond-human knowledge yet good potential  neural architectures.

\begin{figure}[t]
	\centering
		\subfloat[A RNN cell.]{\includegraphics[width=0.28\linewidth]{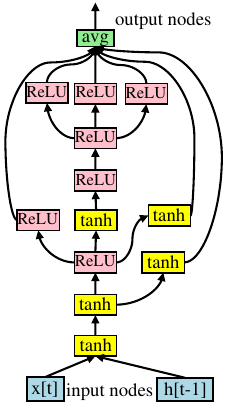}}
	\subfloat[The general model.]{\includegraphics[width=0.36\linewidth]{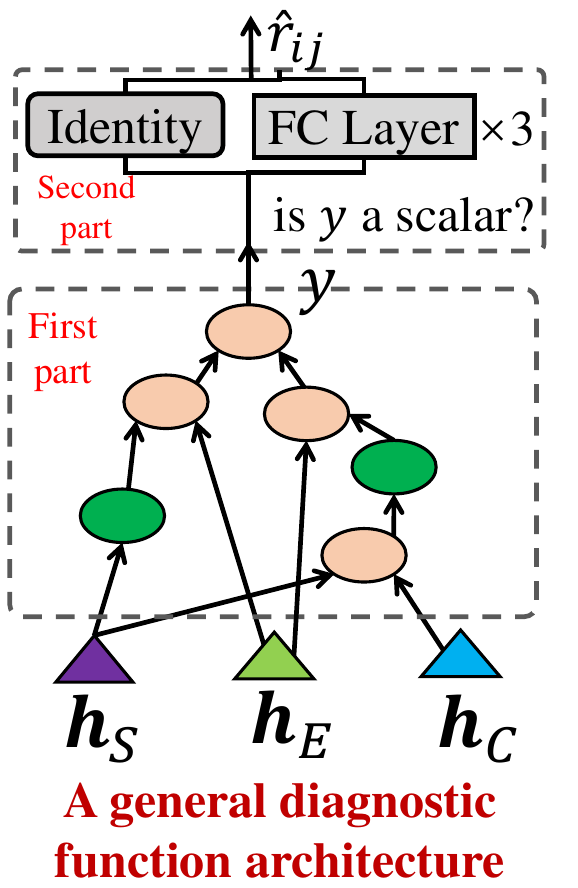}}
	\caption{{Illustration of the proposed  search space (the general diagnostic function architecture). The general model consists of  two parts, and its first part (termed CD cell) holds at most three input nodes  and several computation nodes (unary or binary).}}
	\label{fig:search space}
\end{figure}

To design the search space for CD,  we first observe and summarize existing CD approaches that design novel diagnostic functions.
Then we find that their diagnostic functions combine three types of input vectors in a linear or non-linear manner and finally output a scalar or a vector for the score prediction.
In other words,  the diagnostic function architecture can be seen as a general model that has three input nodes, some internal nodes, and  one output node.
Both its output node and its internal nodes are computation nodes to handle their inputs by their adopted operators.
We can find that the general model  is similar to models under the search space of RNN in NAS~\cite{pham2018efficient,liu2018darts}. 
Fig.~\ref{fig:search space}(a) plots the  RNN cell found by \textit{Efficient Neural
Architecture Search} (ENAS)~\cite{pham2018efficient}, where $x[t]$ and $h[t-1]$ are two input nodes, \textit{avg} is  the output node, and others are computation nodes.

By summarizing previous CD approaches, we collected some operators that be used for computation nodes of the general model.  These operators are divided into two types, i.e., unary and binary operators, which are used to receive one input and two inputs, respectively.
Here  computation nodes (including the output node) in the general model can only handle at most two inputs, which is different from that of RNNs.

 As a result, we take the general model as the proposed search space for CD, where 15 candidate operators in  Table~\ref{tab:operator} can be adopted by each computation node and the following are their descriptions:
\begin{itemize}
	\small
	\item \textbf{Unary operators.} 
	 Each unary operator only takes one input $x$ and returns its output. 	 \textit{FFN\_D}  returns the vectors, \textit{Sum}, \textit{Mean}, and \textit{FFN} return	 the scalar outputs, while the other  eight unary operators  return the outputs having the same shape as their inputs, 
	which contains five  arithmetic operators (i.e., \textit{Neg}, \textit{Abs}, \textit{Inv}, \textit{Square}, and \textit{Sqrt})  and 	 three  activation functions  \textit{Tanh}~\cite{Hu2022Adaptively}, \textit{Sigmoid}~\cite{Uykan2013Fast}, and \textit{Softplus}~\cite{Gu2019Representational}. 
	
	\item 
	\textbf{Binary operators.} 
	Three binary operators considered in the general model:	in addition to addition \textit{Add} and multiplication \textit{Mul}, we further consider a \textit{Concat} operator  to aggregate two input vectors into  one vector.
	Note that the output shapes of \textit{Add} and \textit{Mul}  are determined by the maximal shape of two inputs. 
	For example, when one input  is a scalar $x$ and another input is a  vector $\mathbf{y}\in R^{1\times D}$, 
	the output shape  is same as $\mathbf{y}$ (equal to $1\times D$).
\end{itemize}
Here \textit{FFN} and \textit{Concat} are NN-based operators containing learnable parameters, which make the proposed search space more expressive than that of RNN.

Note that the  general model may output a scalar $y$ or  a vector $\mathbf{y}$,  because the general model may adopt different operators while the output shapes of candidate operators are different.
%Considering the characteristics of different operators, there may output  a scalar $y$ or  a vector $\mathbf{y}$ by the last computation node of the  general model.
To make the prediction process successful, the general model has to execute the following process to get the prediction score of student $s_i$ on exercise $e_j$:
\begin{equation}\label{eqa:second part}
	\hat{r}_{ij}=\left\{
	\begin{aligned}
		&y, \ \mathrm{if}\ y \in R^{1}\\
		&FC_3(FC_2(FC_1(\mathbf{y}))),\ \mathrm{if}\ \mathbf{y} \in R^{1\times D}\\
	\end{aligned}
	\right.,
\end{equation}
 where $FC_1(\cdot)$, $FC_2(\cdot)$, and $FC_3(\cdot)$   are three FC layers with output dimensions $H_1$, $H_2$, and $H_3$, respectively.
The three FC layers are set to hold the monotonicity  property  according to the experiences in~\cite{wang2020neural}.
 By doing so, the probability of a correct response to the exercise is monotonically increasing at any dimension of the student’s knowledge proficiency, which enables FC layers to hold the same  interpretability as the identity operation.

\begin{figure*}[t]
	\centering
	\includegraphics[width=0.65\linewidth]{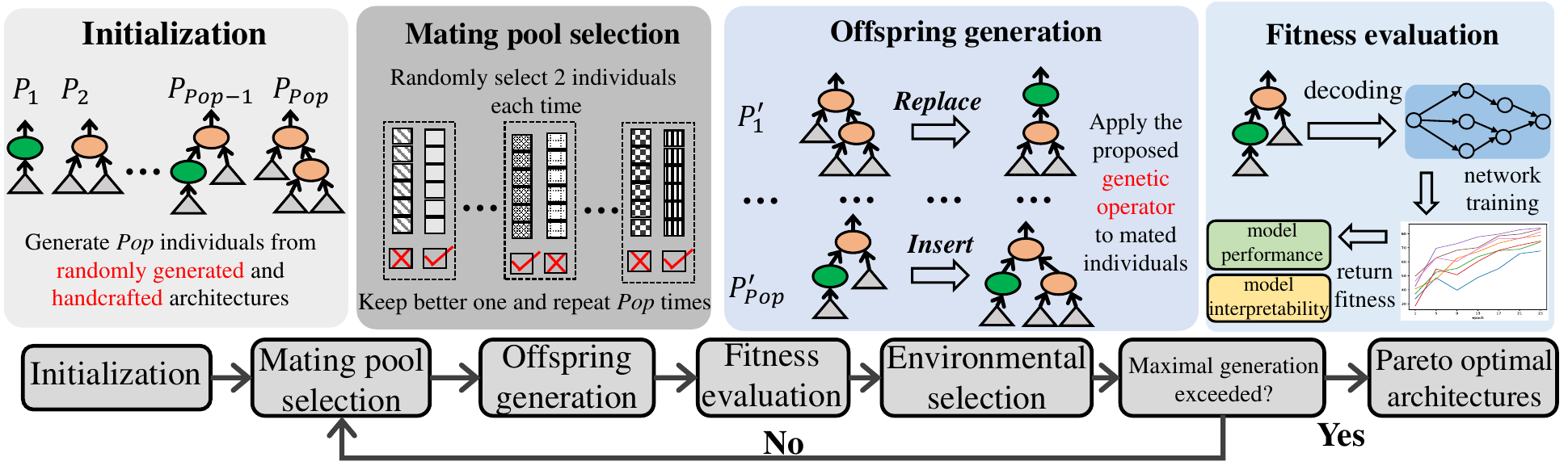}
	\caption{\textbf{Bottom}: the  overall framework of the proposed EMO-NAS-CD; 
		\textbf{Upper}: illustrative examples of four components (population initialization, mating pool selection, offspring generation, and fitness evaluation).}
	\label{fig:framework}
\end{figure*}

For better understanding,  Fig.~\ref{fig:search space}(b) presents  the general  diagnostic function architecture under the proposed  search space.
The general diagnostic function architecture (the general model) contains two parts.
The first part (termed CD cell) is similar to the RNN cell in NAS, 
and the second part is a three-layer FC NN or  an identity operation as shown in (\ref{eqa:second part}).
 The CD cell has  several computation nodes (represented by ovals) and at most three input nodes ($\mathbf{h}_S$, $\mathbf{h}_E$, and $\mathbf{h}_C$, represented by triangles).
 Different from the RNN cell,   its output node is also computation node, 
 and  computation nodes are selected from unary operators (denoted by green nodes) or binary operators (denoted by orange nodes).
After obtaining the CD cell's output $y$, 
either the identity operation or the 
three-layer FC NN  will be  applied to  get the final prediction $\hat{r}_{ij}$.

\begin{figure}[t]
	\centering
	\subfloat[IRT.]{\includegraphics[width=0.216\linewidth]{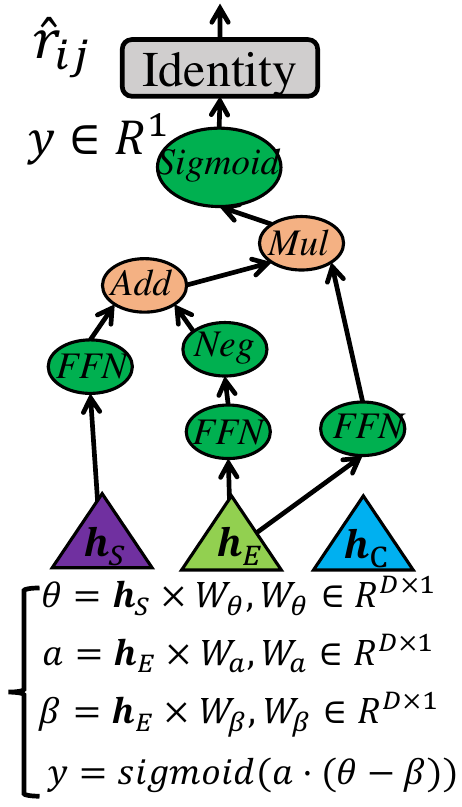}}
	\subfloat[MIRT.]{\includegraphics[width=0.24\linewidth]{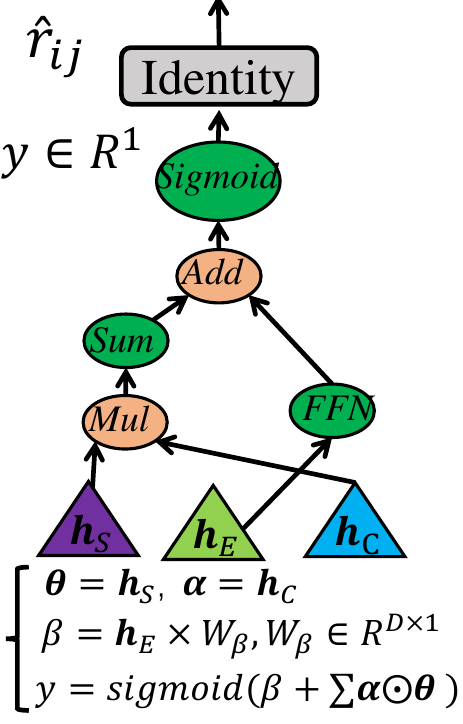}}
	\subfloat[NCD.]{\includegraphics[width=0.318\linewidth]{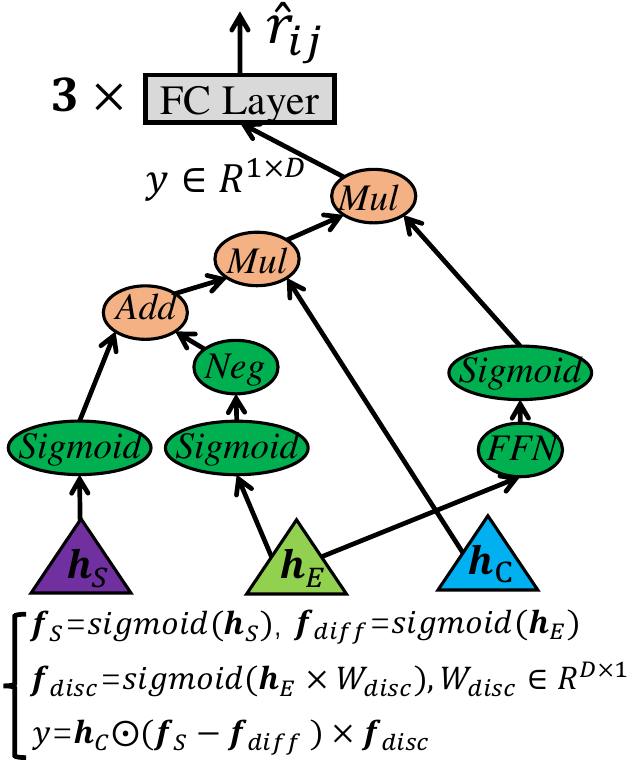}}\\
	\subfloat[MF.]{\includegraphics[width=0.27\linewidth]{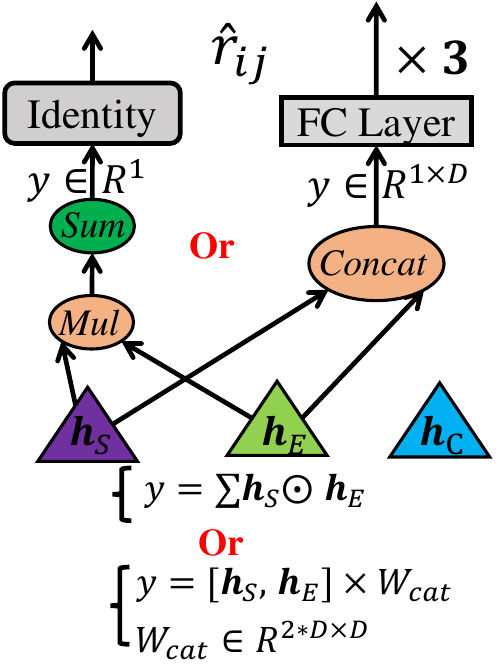}}	
		\subfloat[The CD cell represented by the tree.]{\includegraphics[width=0.65\linewidth]{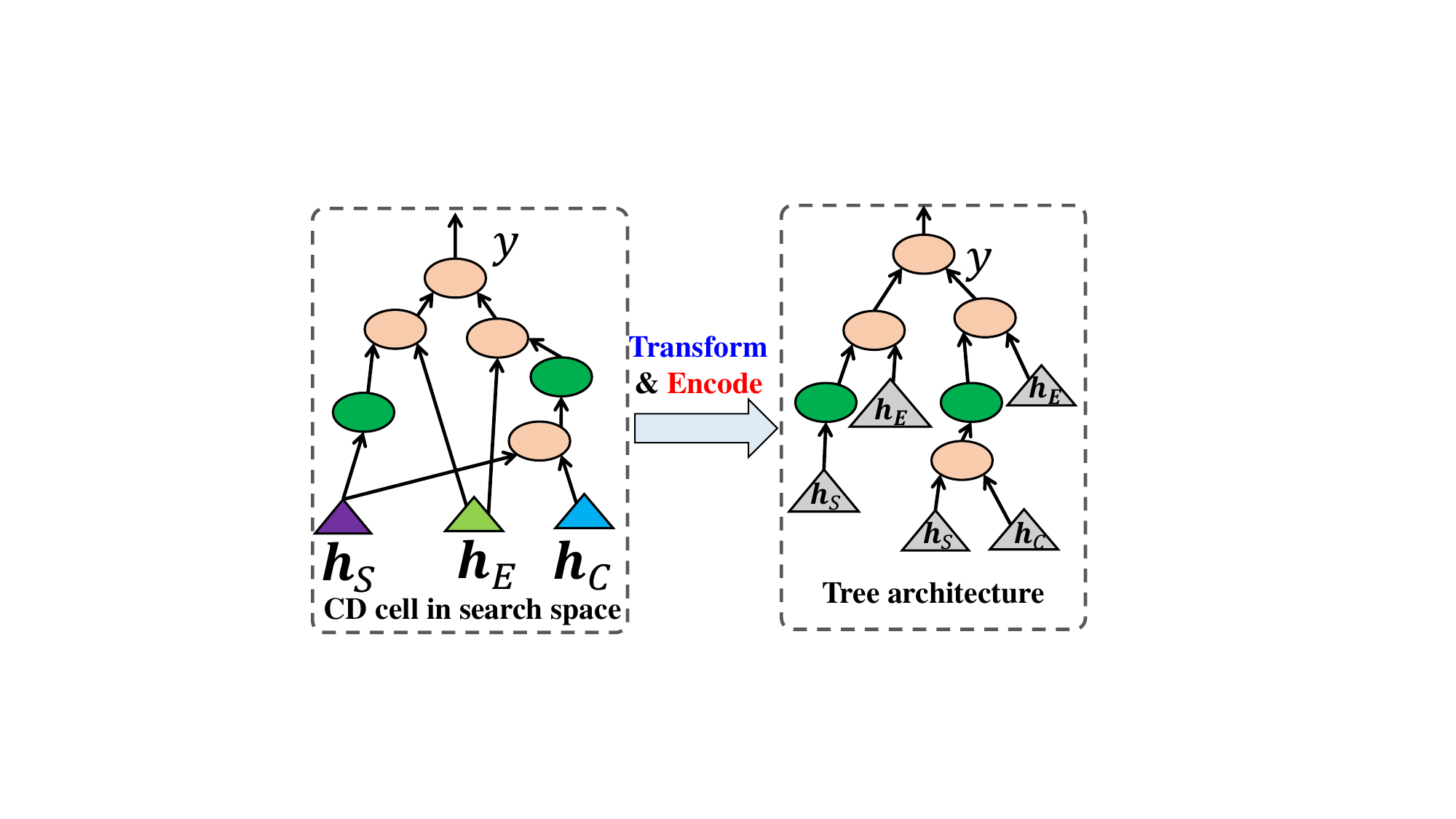}}\\
	\caption{(a), (b), (c)
		\& (d): the diagnostic function architectures of four typical CDMs plotted under the proposed search space; 
	(e): represent the CD cell of the general model by tree encoding.}
	\label{fig:existing CDM arc}
\end{figure}

As stated in~\cite{elsken2019neural,oloulade2021graph}, 
a promising search space should contain not only   a large number of expressive neural architectures but also  as many  existing handcrafted architectures as possible.
To demonstrate the effectiveness of the proposed search space, 
we take four representative CDMs, including IRT, MIRT, MF, and NCD,  as illustrative examples.
Fig.~\ref{fig:existing CDM arc}(a) to Fig.~\ref{fig:existing CDM arc}(d) present their diagnostic function architectures under the proposed search space.
As can be seen, these typical CDMs can be easily represented under  the proposed search space 
by specific computation nodes and selected input nodes.

%说目标
%Three figures to illustrate the interpretability ( 1vs2, 1vs3, 1vs4)
%话三个图说明第二个目标， 深度，宽度，节点数 等等

% 说困难， 是一个树，不同于实数编码整数编码， 比较困难，
%此外， 这个树在构造时存在一定的约束，这让搜索更加困难。

\section{The Proposed EMO-NAS-CD}
This section will first present the proposed EMO-NAS-CD framework,
and  then sequentially give individual representation, objectives, and a tailored genetic operation.
Finally,  other  details are introduced.

\subsection{Overall Framework of EMO-NAS-CD}
% main idea
The main idea of the proposed EMO-NAS-CD is to search high-performance  diagnostic function architectures holding high interpretability under the devised search space.
To this end,  we aim to solve the NAS-CD task by optimizing a multi-objective optimization problem (MOP), which has  two objectives:  model performance and model interpretability.
To avoid the difficulties of using vector-based encoding for the devised search space (e.g., variable-length encoding problem),
we propose  MOGP (a popular type of MOEAs~\cite{bartz2014evolutionary}) to solve the MOP by transforming architectures into tree architectures and encoding them by trees, 
 because genetic programming (GP)~\cite{Shao2014Feature,chen2017feature,bi2020genetic} can  solve  tree-encoding-based problems well.
The devised MOGP follows the framework of NSGA-II~\cite{deb2002fast}, and we devise  an effective genetic operation  and   a population initialization strategy for the MOGP.
As can be seen that the proposed EMO-NAS-CD is  a MOGP-based NAS approach for CD.

\iffalse
Based on the  classical NSGA-II~\cite{deb2002fast}, 
the main idea of the proposed EMO-NAS-CD is to search effective diagnostic function architectures holding high interpretability by maximizing the objectives of model performance and model interpretability.
Instead of using \XG{vector-based encoding} for each architecture in the proposed search space, 
we first transform each  architecture  into its corresponding tree architecture,
and then encode it by the tree-based representation, 
which avoids some difficulties of \XG{vector-based encoding} (e.g., variable-length encoding difficulty) in general  MOEAs
and is easier to be optimized by GP~\cite{chen2017feature,bi2020genetic}.
Moreover, we devise  an effective genetic operation inspired by GP and   a population initialization strategy for the  proposed EMO-NAS-CD.
As a result, the proposed EMO-NAS-CD is a MOGP-based NAS approach for CD.
\fi

% framework

\begin{algorithm}[t!]
	\footnotesize
	\caption{Main Steps of EMO-NAS-CD}
	\label{algorithm: EMO-NAS-CD}

	\begin{algorithmic}[1]		
		\REQUIRE
		{$Gen$: Maximum number of generations;
			$Pop$: Population size;
			$Node_{range}$: Range of the number of nodes in initialization;}
		\ENSURE
		$\mathbf{P_{non}}$: Non-dominated individuals;
		\STATE $\mathbf{P} \leftarrow Initialization(Pop,Node_{range})$;

		\STATE  Evaluate each individual in $\mathbf{P}$ by training its corresponding architecture for $Num_{E}$ epochs and obtain the objective \\values;
		
		\FOR{$g=1$ to $Gen$}
		\STATE $\mathbf{P}' \leftarrow$ Select $Pop$ parent individuals from Population $\mathbf{P}$; \\   \% \texttt{Mating pool selection}
		
		\STATE $\mathbf{Q} \leftarrow {\rm Genetic\ Operation}(\mathbf{P}')$; \% \texttt{Algorithm~\ref{alg:genetic_operator}}
		
		\STATE  Evaluate each individual in $\mathbf{Q}$ by training its corresponding architecture for $Num_{E}$ epochs and obtain the  objective values;

		\STATE $\mathbf{P} \leftarrow {\rm Environment\ Selection}(\mathbf{P} \cup \mathbf{Q})$;	
		
		\ENDFOR
		\STATE		Select non-dominated individuals $\mathbf{P_{non}}$ from $\mathbf{P}$ as the output
		
		\RETURN $\mathbf{P_{non}}$ ;
	\end{algorithmic}
\end{algorithm}

The overall framework of the proposed EMO-NAS-CD is summarized in Fig.~\ref{fig:framework}, which is mainly composed of five steps.
Firstly, a population initialization strategy (in Section~\ref{sec:initial}) is employed to randomly generate $Pop$ individuals as population $\mathbf{P}$.
% where partial  architectures are  initialized from handcrafted architectures' variants to provide a fast convergence speed to avoid  completely evolving from scratch.
Second,  the standard binary tournament selection is employed to select individuals for  getting the mating pool $\mathbf{P}'$.
Next, a novel genetic operation  is applied to  $\mathbf{P}'$ to generate offspring individuals and form the offspring population $\mathbf{Q}$.
Fourth, train the   architecture of each individual of $\mathbf{Q}$ for a certain number of ($Num_{E}$) epochs to compute its objective values.
Fifth, the environmental selection in NSGA-II~\cite{deb2002fast} will be employed to 
identify and maintain the individuals that hold better objective values from the  union of population $\mathbf{P}$ and offspring population $\mathbf{Q}$.
The second to the fifth step will be repeated until the maximal number of generation $Gen$ is exceeded,
then the non-dominated individuals will  finally be output.
For details, Algorithm~\ref{algorithm: EMO-NAS-CD} also summarizes the main procedures of the proposed EMO-NAS-CD.

It is worth noting that there exist some individuals during the whole optimization process,
whose  neural architectures achieve terrible performance, nearly close to random performance.
The reason  behind this is that these architectures will encounter the gradient explosion problem when they continuously use some operations (e.g., \textit{Square}, \textit{Tanh}, and \textit{Softplus}), which makes it difficult for  general training paradigms to train them well.
To solve this problem, in the individual evaluation, 
we adopt a simple early-stopping strategy~\cite{elsken2019neural} to stop the training of a neural architecture  if its performance does not improve for several epochs.

\subsection{Individual Representation}
To represent architectures in the proposed search space, 
vector-based encoding is naturally our first choice because of its high popularity in many real-world optimization problems.
Suppose the vector-based encoding for $i$-th computation node of an architecture is  $n_i=\{link_1, link_2,Op\}$, 
where ${link}_1$ and ${link}_2$ denote node $n_i$ receiving  which nodes' outputs  and $Op$ denotes which operator is adopted, 
and then each architecture is represented  by a set of nodes $\{n_i| 1\le i \le num_{c}\}$ ($num_{c}$ denotes the number of computation nodes). 

However, as shown in Fig.~\ref{fig:search space}(b), the architectures in the proposed search space are variable. Thus it is difficult and unsuitable to represent architectures by vector-based encoding due to two challenges.
The first challenge is that $num_{c}$ is not fixed but variable, 
and thus the vector-based encoding of each architecture is variable-length, 
which is difficult to solve by general MOEAs~\cite{liu2021survey}.
Secondly, different from the output node of the RNN cell, 
the output node in the proposed search space is  a computation node and receives at most two inputs.
This poses a decision constraint in using vector-based encoding as individual representation and thus is also difficult to solve.

\iffalse
It can be found from Fig.~\ref{fig:search space}(b) that there are two challenges for \XG{vector-based encoding} in general MOEAs to represent architectures in the proposed search space.
Suppose the \XG{vector-based encoding} for $i$-th computation node of an architecture is  $n_i=\{link_1, link_2,Op\}$, 
where ${link}_1$ and ${link}_2$ denote node $n_i$ receiving  which nodes' outputs  and $Op$ denotes which operator is adopted, 
and then each architecture is represented  by a set of nodes $\{n_i| 1\le i \le num_{c}\}$ ($num_{c}$ denotes the number of computation nodes). 
The first challenge is that $num_{c}$ is not fixed but variable and thus the \XG{vector-based encoding} of each architecture is variable-length, which is difficult to solve by general MOEAs~\cite{liu2021survey}.
Secondly, different from the output node of the RNN search space, the output node in the proposed search space is  a computation node and thus receives at most two inputs, 
which poses a decision constraint in using \XG{vector-based encoding} as individual representation and thus is also difficult to solve.
\fi

To avoid the above issues, we propose to  utilize tree-based representation to encode architectures in our proposed search space, 
and  we propose MOGP to solve the MOP  to search novel CDMs because of the superiority of GP in solving tree-encoding-based optimization problems~\cite{bi2020genetic}.
For this aim,  we have to transform the architectures under the proposed search space into their  
corresponding single-root tree architecture.
Fig.~\ref{fig:existing CDM arc} (e) gives the transform process by taking the general model as an illustrative example: 
the input nodes  are seen as \textbf{the leaf nodes} of the tree architecture,  the output node is equal to \textbf{the root node}, 
and the whole tree architecture can be seen as a single-root binary computation tree, 
where the obtained tree architecture is similar to the Koza-like tree  in GP~\cite{maitre2010easea}.
Based on the tree-based representation,  the proposed MOGP can effectively search diagnostic function architectures but still needs the assistance of some tailored strategies, such as genetic operations and initialization strategies.

\iffalse
we observe that the general model  can be transformed into a corresponding single-root tree architecture. 
Fig.~\ref{fig:existing CDM arc} (e) gives the transform process: 
the input nodes  are seen as \textbf{the leaf nodes} of the tree architecture,  the output node is equal to\textbf{ the root node}, 
and the whole tree architecture can be seen as a single-root binary computation tree, 
where the obtained tree architecture is similar to the Koza-like tree  in GP~\cite{maitre2010easea}.
Considering the superiority of GP in solving \XG{tree-encoding-based optimization problems~\cite{bi2020genetic}}, 
 we adopt tree-based representation to encode  architectures in our search space, 
 and thus the proposed MOEA turns out to be a MOGP, which needs effective tailored genetic operations.
\fi

\subsection{Objectives}

To make the searched architectures hold good  performance and high interpretability, 
the proposed MOGP is to optimize the following MOP:
\begin{equation}\label{eqa:MOP}
	\max_{\mathcal{A}} F(\mathcal{A})=\left\{
	\begin{aligned}
		&f_1(\mathcal{A}) =  AUC(\mathcal{A},D_{val}) \\
		&f_2(\mathcal{A}) = {\rm model\ interpretability(\mathcal{A}) } \\
	\end{aligned}
	\right.,
\end{equation}
where $\mathcal{A}$ denotes the candidate  architecture to be optimized. 
$f_1(\mathcal{A})$ represents the  AUC (\textit{Area Under an ROC Curve}) value~\cite{bradley1997use}  of  $\mathcal{A}$ (i.e., model performance) on validation dataset $D_{val}$. 
$f_2(\mathcal{A})$ represents the model interpretability of architecture $\mathcal{A}$, 
since an architecture holding high model interpretability is preferred for CD.

To obtain reasonable $f_2(\mathcal{A})$, 
an intuitive idea is to  compute the model complexity by counting
how many computation  nodes and leaf nodes are in $\mathcal{A}$. 
But it is not reasonable~\cite{virgolin2020learning} to some extent 
	since much research~\cite{zhang2021survey} indicates that the model depth plays the most important role in the model interpretability. 
	Besides, some  research on interpretable trees~\cite{hein2018interpretable,virgolin2020learning} further indicates that  binary operators commonly provide better interpretability than unary operators.	More importantly, recent CDMs prefer introducing extra inputs and more feature fusions in the models because it is easier to interpret the model performance~\cite{ma2022knowledge,zhou2021modeling}. 
	This implies that more inputs in CDMs represent higher interpretability, 
	further indicating that  binary operators are more important than unary  ones since  binary operators  	will introduce more inputs.

\begin{figure}[t]
	\centering
	\subfloat[$\mathcal{A}_1$.]{\includegraphics[width=0.12\linewidth]{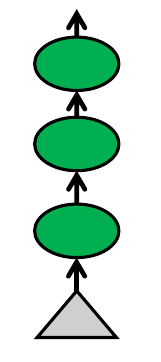}}
	\subfloat[$\mathcal{A}_2$.]{\includegraphics[width=0.144\linewidth]{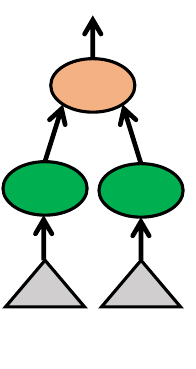}}
	\subfloat[$\mathcal{A}_3$.]{\includegraphics[width=0.23\linewidth]{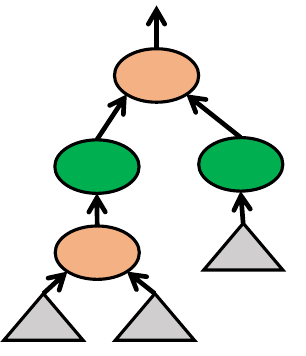}}
	\subfloat[$\mathcal{A}_4$.]{\includegraphics[width=0.24\linewidth]{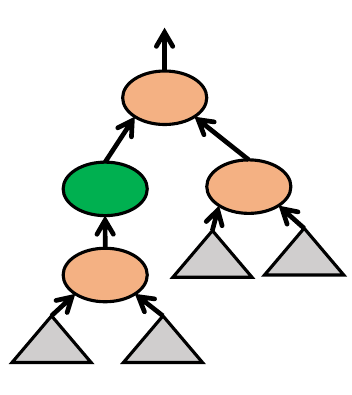}}
	\subfloat[$\mathcal{A}_5$.]{\includegraphics[width=0.24\linewidth]{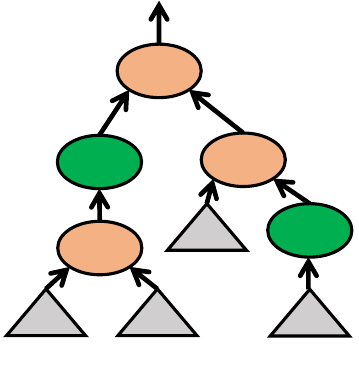}}\\
	\caption{An illustration of the  model interpretability objective by five  diagnostic function architectures including $\mathcal{A}_1$, $\mathcal{A}_2$, $\mathcal{A}_3$, $\mathcal{A}_4$, and $\mathcal{A}_5$.} 
		%where $\mathcal{A}_2$ has the best  model interpretability followed by 		$\mathcal{A}_4$, $\mathcal{A}_5$, $\mathcal{A}_3$, and $\mathcal{A}_1$. }
	\label{fig:objectives}
\end{figure}

However, the model complexity  that counts the number of nodes in $\mathcal{A}$ can not reflect the above fact.
As shown in Fig.~\ref{fig:objectives}, despite more nodes, we think  $\mathcal{A}_2$ holds better interpretability than $\mathcal{A}_1$  due to a smaller depth.
Due to  larger breadth (more inputs), 
$\mathcal{A}_4$ and $\mathcal{A}_3$ should be better than $\mathcal{A}_1$ but worse than  $\mathcal{A}_2$. 
$\mathcal{A}_5$ should be better than $\mathcal{A}_3$ but worse than $\mathcal{A}_4$ due to containing more nodes.

\iffalse
Even compared to $\mathcal{A}_3$  having the same depth as $\mathcal{A}_1$,
$\mathcal{A}_1$ is worse than  $\mathcal{A}_3$ because $\mathcal{A}_3$ holds a larger breadth than $\mathcal{A}_1$, 
where the tree breadth  is equal to the number of leaf nodes.
As can be seen from the comparisons among  $\mathcal{A}_1$, $\mathcal{A}_3$, and $\mathcal{A}_4$,   
a tree holding a larger breadth means  more binary operators  contained in the tree,  
and thus indicates the tree holds higher interpretability.
Besides,  $\mathcal{A}_4$  holds  higher interpretability than $\mathcal{A}_5$ 
since $\mathcal{A}_4$ has  fewer computation nodes.
\fi

With  the above considerations, we characterize the model interpretability of  architecture $\mathcal{A}$ by its tree's depth, breadth, and  computation node number. 
The model interpretability of $\mathcal{A}$ is first determined by the tree depth $depth$,
then by the tree breadth $breadth$ (equal to the number of leaf nodes), and finally  by  the number of computation nodes $num_c$. As a consequence, the $f_2(\mathcal{A})$ can be computed by
\begin{equation}\label{eqa:f2}
	f_2(\mathcal{A}) = (1-\frac{depth-1}{10})+\frac{breadth}{200}+(0.001- \frac{num_c}{20000}),
\end{equation}
where  we make the depths of all architectures  less than 10 in this paper to hold high model interpretability 
and thus $f_2(\mathcal{A})\in (0,1)$ has five decimal places.
The first decimal place is determined by $depth$, 
The second and third  decimal places are  determined by $breadth$, and 
the remaining decimal places are  determined by $num_c$.
Note that three parameters (10, 200, 2000) are  empirically set and can be other choices, 
which will not affect the proposed approach's result as long as two criteria are met.
Firstly, the decimal place(s) determined by $depth$, $breadth$, and $num_c$ do not affect each other;
secondly, the decimal place(s) determined by $depth$ is most important,  followed by $breadth$, and finally $num_c$.

In Fig.~\ref{fig:objectives}, the depths of five architectures are $3$, $2$, $3$, $3$, and $3$,
their breadths are $1$, $2$, $3$, $4$, and $4$, and their computation node numbers are $3$, $3$, $4$, $4$, and $5$.
According to (\ref{eqa:f2}), 
their second objective values are $0.80585$, $0.91085$, $0.81580$, $0.82080$, and $0.82075$, respectively,  which are consistent with our consideration.

\subsection{Genetic Operation}\label{sec:genetic}

For effective offspring generation in the proposed MOGP, we propose an effective genetic operation based on four  sub-genetic operations that modified and inspired from GP~\cite{Shao2014Feature,chen2017feature,bi2020genetic}.

The following  introduces four modified sub-genetic operations: \textit{Exchange}, \textit{Delete}, \textit{Replace}, and \textit{Insert}. 
\begin{itemize}
	\small
	\item \textit{Exchange.} Given two  individuals, $\mathbf{P}'_{1}$ and $\mathbf{P}'_{2}$,  
	randomly select two sub-trees, $t_1$ and $t_2$,
 from the trees corresponding to two individuals,  respectively,
 and then exchange two sub-trees to generate two new trees and form two offspring individuals, $\mathbf{O}_{1}$ and   $\mathbf{O}_{2}$. (The root nodes  will  not be selected.)

	\item \textit{Delete.} Given a parent individual $\mathbf{P}'_{1}$,
	randomly select a computation node from the tree corresponding to $\mathbf{P}'_{1}$.
	To delete this node, one of the left and right child trees of this node will be   randomly connected to its parent node (if exists).  The newly generated tree can  form  the offspring individual $\mathbf{O}_{1}$.
	
	\item \textit{Replace.} For the tree corresponding to individual $\mathbf{P}'_{1}$, 
	randomly select a node to be replaced and replace the node's operator by a new operator randomly sampled from Table~\ref{tab:operator}.
	If the original operator is unary but the sampled operator is binary,
	 a new leaf node will be generated and connected to this node as its child tree, 
	 where the  new leaf node  is randomly sampled from $\{\mathbf{h}_S, \mathbf{h}_E, \mathbf{h}_C\}$.
	 If the original operator is binary but the sampled operator is unary,
	 only one of the left and right child trees of this node  will be kept. 
	 As a result,  offspring individual $\mathbf{O}_{1}$ can be obtained based on the revised tree.  
	
	\item \textit{Insert.} {A new operator is first randomly sampled from the predefined operators,}
	and a computation node is  randomly selected from    individual $\mathbf{P}'_{1}$.
Then, the sampled  operator is inserted between this node and its parent node (if exists) as a new computation node.
If the sampled  operator is binary, an additional leaf node will be randomly sampled from $\{\mathbf{h}_S, \mathbf{h}_E, \mathbf{h}_C\}$ and added to the new computation node as its child tree.
Finally,  offspring individual $\mathbf{O}_{1}$ will be generated.	
\end{itemize}

\begin{figure}[t]
	\centering
	\subfloat[\textit{Exchange}.]{\includegraphics[width=0.425\linewidth]{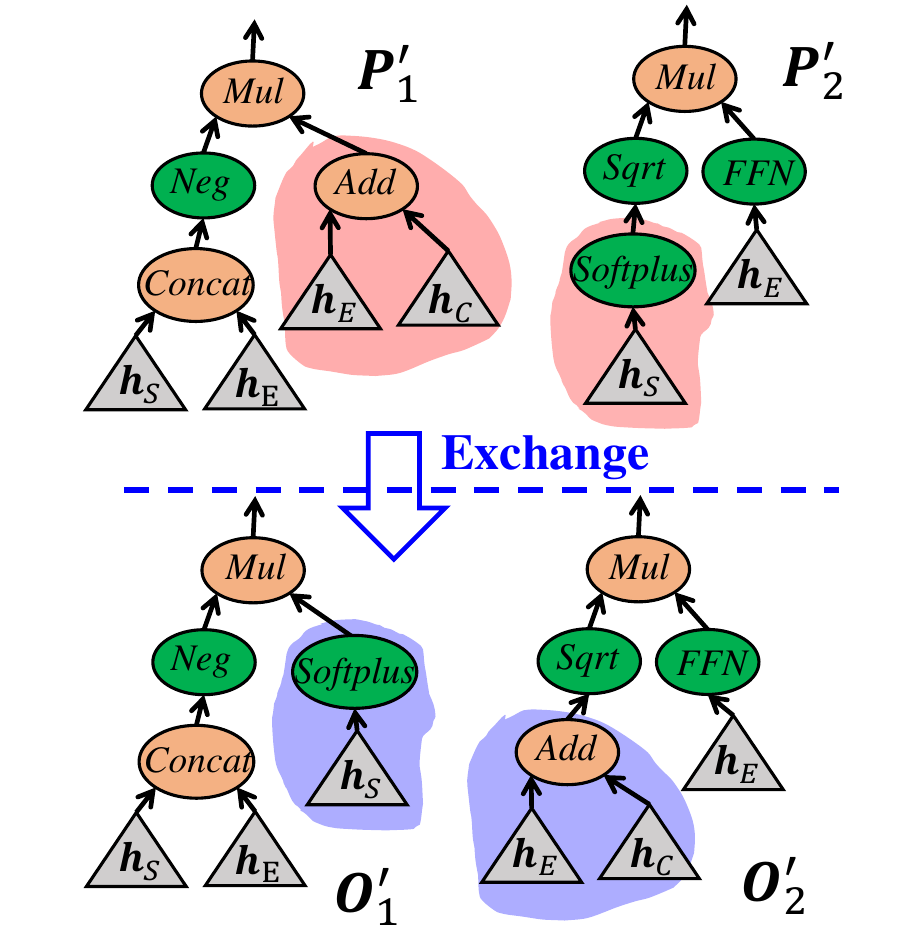}}
	\subfloat[\textit{Delete}.]{\includegraphics[width=0.204\linewidth]{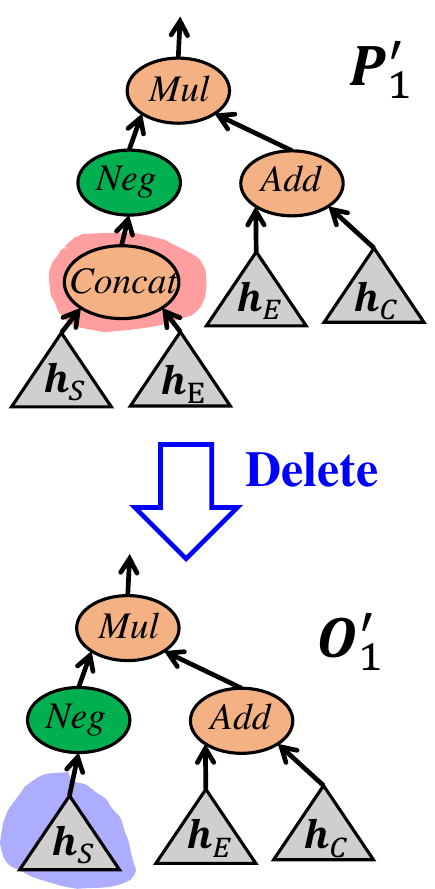}}
	\subfloat[\textit{Replace}.]{\includegraphics[width=0.19\linewidth]{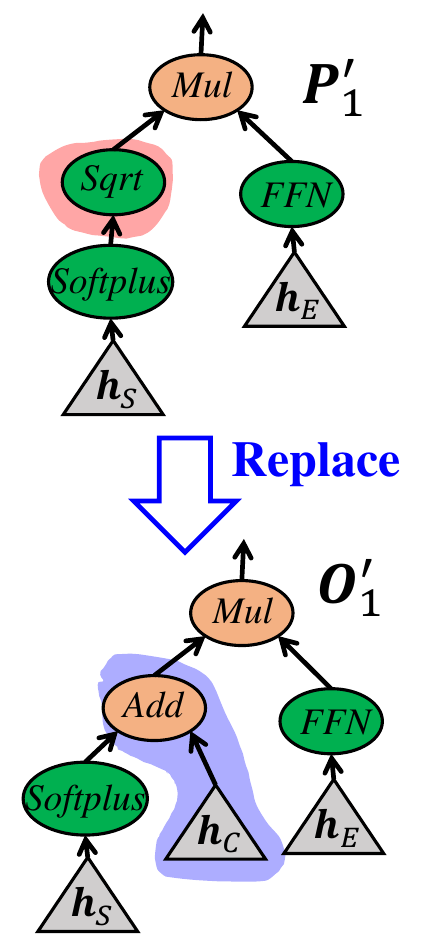}}
	\subfloat[\textit{Insert}.]{\includegraphics[width=0.20\linewidth]{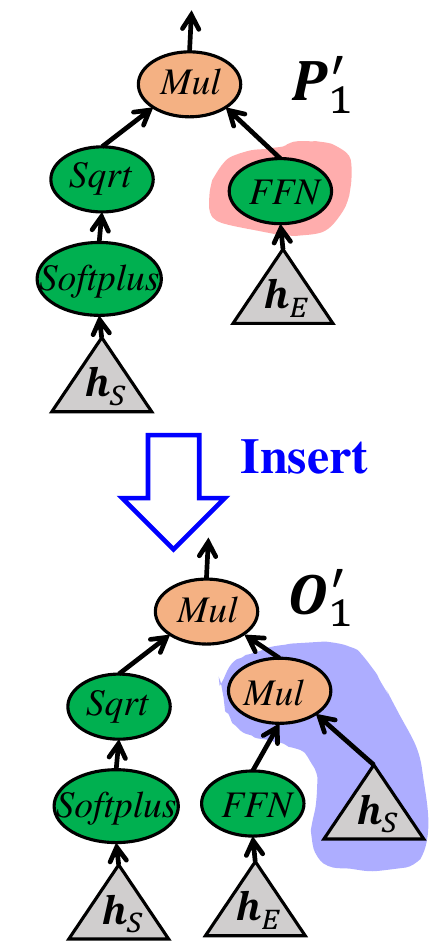}}
	\caption{Illustrative examples of four sub-genetic operations, where the second parts are not plotted. }
	\label{fig:genetic operator}
\end{figure}

{Note that the root node will not be involved  in \textit{Exchange} since the \textit{Exchange}  operation will be meaningless or ineffective if  root nodes are selected. }
For a better understanding of the above operations, 
Fig.~\ref{fig:genetic operator} gives  some illustrative examples of generating offspring individuals.
The pink area denotes the selected computation nodes (or corresponding sub-trees) needed to be handled, 
and the light purple area represents   the executed changes.

As can be seen,  \textit{Exchange}   will  lead to big modifications  between generated individuals and   corresponding parent individuals, while other operations commonly lead to small modifications.
Therefore, the \textit{Exchange} operation can be used for exploration, and others can be used for exploitation~\cite{vcrepinvsek2013exploration}. % ~\cite{saka2016metaheuristics}
Equipped with  four sub-genetic operations, we empirically combine them to  form our proposed  genetic operation, whose basic procedures are summarized in Algorithm~\ref{alg:genetic_operator}.
Four operations are called four sub-genetic operations because they can constitute  many other genetic operations when adopting different combination manners.
In Algorithm~\ref{alg:genetic_operator}, {two individuals $\mathbf{P}'_{i}$ ($i$-th individual in $\mathbf{P}'$) and $\mathbf{P}'_{i+1}$ are first selected from the mating pool  $\mathbf{P}'$, 
and  the numbers of computation nodes in the two individuals are computed as $num_c^i$ and 
$num_c^{i+1}$ (Lines 3-4). }
Second, randomly sample an integer $rand$ from $\{1,2,3,4\}$ if both  $num_c^i$ and $num_c^{i+1}$ are not smaller than 2, otherwise  randomly sample $rand$ from $\{3,4\}$. 
Numbers $1$, $2$, $3$, and $4$  correspond to \textit{Exchange}, \textit{Delete}, \textit{Replace}, and \textit{Insert} , respectively (Lines 5-9).
This is because \textit{Exchange} and \textit{Delete} will  be ineffective, even meaningless, 
if there is only one computation node in the individual.
Third,  the sub-genetic operation corresponding to $rand$ will be applied to  $\mathbf{P}'_{i}$ and $\mathbf{P}'_{i+1}$  to generate offspring individuals $\mathbf{O}_1$ and  $\mathbf{O}_{2}$ (Lines 10-14).
Next, the obtained  $\mathbf{O}_1$ and  $\mathbf{O}_{2}$  will be added to the offspring population $Off$ (Line 15).
The first to the fourth step will be repeated until all offspring individuals are generated.
After that, an individual repair strategy in Section~\ref{sec:repairing} is used to
make offspring individuals feasible
since there exist some constraints for some operators in computation nodes of trees (Line 17).
For example,  \textit{Sum}, \textit{Mean}, \textit{FFN}, and \textit{Concat} only receive vectors as their inputs. 
Finally, the obtained offspring population $Off$ is output.

\begin{algorithm}[t]
		\footnotesize
	\caption{{Genetic Operation}}	
	\begin{algorithmic}[1]		
		\REQUIRE
		{$\mathbf{P}'$:Mating pool, 
			$Pop$: Population size		
		}
		\ENSURE
		$Off$: Offspring ;
		\STATE $Off \leftarrow \emptyset$;
		\FOR{$i=1$ to $Pop$}
		
		\STATE	$\mathbf{P}'_{i}, \mathbf{P}'_{i+1}\leftarrow$ Sequentially select two individuals from $\mathbf{P}'$;	
		\STATE $num_c^i, num_c^{i+1} \leftarrow$ Get the numbers of computation nodes in  $\mathbf{P}'_{i}$ and $\mathbf{P}'_{i+1}$;
			
		\IF{ $ num_c^i\geq 2$ and $ num_c^{i+1}\geq 2$}		
		\STATE		$rand\leftarrow$ Randomly sample an integer from $\{1,2,3,4\}$; 		
		\ELSE
		\STATE 			$rand\leftarrow$ Randomly sample an integer from $\{3,4\}$; 
		\ENDIF

		\IF {$rand = 1$}
		\STATE $\mathbf{O}_1, \mathbf{O}_{2}\leftarrow $ Apply \textit{Exchange}  to $\mathbf{P}'_{i}$ and $\mathbf{P}'_{i+1}$;
		\ELSIF{$rand = 2/3/4$}
		\STATE $\mathbf{O}_1, \mathbf{O}_{2}\leftarrow $ Apply \textit{Delete}/\textit{Replace}/\textit{Insert} to $\mathbf{P}'_{i}$ and $\mathbf{P}'_{i+1}$;
		\ENDIF
		
		\STATE 	$Off\leftarrow Off \cup \mathbf{O}_1 \cup\mathbf{O}_{2}$, $i\leftarrow i+2$;		
		\ENDFOR
		
		\STATE Individual repairing for $Off$; \% \texttt{Section~\ref{sec:repairing}}

		\RETURN $Off$;
	\end{algorithmic}
\label{alg:genetic_operator}
\end{algorithm}

\subsection{Related Details}
In the mating pool selection  of  EMO-NAS-CD,
two individuals are first randomly selected each time, 
and then their non-dominated front sizes and crowding distance values are compared to   keep the better one. 
 The computation of non-dominated front size and crowding distance for each individual is the same as for NSGA-II~\cite{deb2002fast}.
Due to the simple topologies of tree architectures, this will generate many duplicated individuals.
To address this issue, a simple archive stores the individuals that have appeared and identifies whether a newly generated individual has already occurred.
In addition, there are the population initialization strategy and the individual repairing strategy in the proposed approach.

\subsubsection{Population Initialization}\label{sec:initial}
Instead of evolving architectures entirely  from scratch~\cite{zoph2016neural,liu2021survey},
we aim to introduce  prior knowledge about existing CDMs' diagnostic functions into the  search process.
%By doing so,  
% the convergence of the evolution can be accelerated to achieve  better final convergence to some extent 
% according to the experiences in.

{To this end, one half of the individuals in the population are generated from four existing CDMs (IRT, MIRT, MF, and NCD) by applying the proposed  genetic operation.}
To maintain the diversity of the population and avoid getting trapped into local optima, 
another half of individuals are randomly generated from scratch. 
Here, we utilize a hyperparameter  $Node_{range} = \{node_{h1}, node_{h2}\}$ to limit the computation node number 
sampled in each randomly generated individual.
Here $node_{h1}$ and $node_{h2}$  refer to the lower and upper bounds of the number of generated nodes.

%In addition to randomly sampling computation nodes from all candidate operators, 
%we further employ a probability $prob =\frac{i}{Pop}$  to sample the computation nodes for $i$-th individual $\mathbf{P}_i$ according to the experiences in~\cite{yang2022accelerating,tian2019evoFuzz}.
%For each node in $\mathbf{P}_i$,  an operator is randomly sampled only from unary operators or only from binary operators, which is determined by the probability $prob$.
%By doing so, the initial population will maintain good diversity.

%As a result, for each individual, we first get its number of computation nodes by sampling from $Node_{range}$,
%then randomly select the manner of random node sampling  or the manner of probability $prob$-based node sampling. 

\subsubsection{{Individual Repair}}\label{sec:repairing}
Most operators in Table~\ref{tab:operator} can be applied to the input with any shape, 
except for \textit{Sum}, \textit{Mean},   \textit{FFN}, and \textit{Concat}, which can only receive one-dimensional vectors as their inputs.
 The first three operators are specially used to extract a high-level scalar feature  from vectors, while \textit{Concat} is specially used to concatenate and map two vectors to one vector.

Therefore, one generated   individual is infeasible and needs repairing if its contained nodes  are equipped with the above four operators but take scalar  inputs (termed infeasible nodes).
To tackle this issue,  we first execute the post-order traversal for each individual to check whether each node is feasible and then directly replace the operator of the infeasible node  with other unary operators or other binary operators (e.g, replace \textit{Concat} by  \textit{Add}, and replace \textit{Mean} by \textit{Neg}).
%~\cite{andersson1990construction}

\subsubsection{{Complexity Analysis}}
{The time complexity of the proposed EMO-NAS-CD is mainly determined by two components, i.e., the training  of each architecture and the optimization process of NSGA-II. Suppose the size of a training dataset is $|D_{train}|$, 
	%and the number of training epochs is $Num_E$, 
 the  time complexity of training each architecture~\cite{mu2016stochastic} is $O(Num_E\times  |D_{train}| \times D)$, and the time complexity of  one generation of NSGA-II is $O(Pop^2)$~\cite{deb2002fast}. Therefore, the overall time complexity of EMO-NAS-CD is
 $O(Pop\times Gen \times Num_E\times  |D_{train}| \times D) +O(Pop^2\times Gen)$. Since $  Num_E\times  |D_{train}| \times D \gg Pop\times Gen$, the time complexity of EMO-NAS-CD can be regarded as $ O(Pop\times Gen \times Num_E\times  |D_{train}| \times D)$.}
 
On the other hand, its space complexity is mainly determined by  the population and the offspring population, and each population has $Pop$ individuals encoded by trees.  
	Suppose the average number of computation nodes in the trees  is $AvgNum$, 
	the space complexity of an individual is $O(AvgNum*3)$ since each node needs three numbers to specify its  operation and two subtrees.
	As a result,  the whole space complexity of EMO-NAS-CD is $O(AvgNum*3\times Pop \times 2)$, i.e., $O(AvgNum\times Pop \times 6)$.
	
	\iffalse
	of EMO-NAS-CD is determined by hidden vectors with the size of $1\times D$ in  each architecture, where the number of hidden vectors is determined by the number of  contained  computation nodes and three leaf nodes.
 Suppose the average number of computation nodes in each architecture is $AvgNum_{node}$, 
 the space complexity of EMO-NAS-CD is $O(Pop\times Gen \times(AvgNum_{node}+3)\times D)$.
\fi

\section{Experiments}
%主实验
% 把f2切换为节点个数后，对比结构和性能
%

\begin{table}[t]
		\renewcommand{\arraystretch}{1}
		\caption{{Statistics of two popular education datasets.}}
			\centering
	\setlength{\tabcolsep}{2.0mm}{
		\begin{tabular}{ccc}
			\toprule
			\textbf{Dataset}& ASSISTments~\cite{feng2009addressing}$^{\textcircled{1}}$& SLP~\cite{slp2021edudata}$^{\textcircled{2}}$ \\
			\midrule
			\# Students& 4,163  & 1,499 \\
			\# Concepts& 123 & 34\\
			\# Exercises&17,746 & 907 \\
			\# Response logs& 324,572 & 57,244\\
			\bottomrule
	\end{tabular}}

%	\begin{tablenotes}
		\centering
%\item $\textcircled{1}$ \url{https://sites.google.com/site/assistmentsdata/} 
%\item $\textcircled{2}$ \url{https://aic-fe.bnu.edu.cn/en/data/index.html}
%\end{tablenotes}	
	\label{tab:dataset}
\end{table}

\subsection{Experimental Settings}

\begin{table*}[t]
	\centering
	\renewcommand{\arraystretch}{0.0}
	
	\caption{The prediction performance comparison between  the architectures found by the proposed EMO-NAS-CD (one run) and comparison CDMs in terms of ACC, RMSE, and AUC values (averaged on 30 runs), obtained on the test datasets of  ASSISTments and SLP. Best results in each column (excluding $\overline{\mathbf{A1}}$-$\overline{\mathbf{A7}}$ and $\overline{\mathbf{S1}}$-$\overline{\mathbf{S7}}$) are highlighted. }
	%\caption{The Prediction Performance Comparison of The Architectures Found by The Proposed EMO-NAS-CD and The State-of-the-art CDMs in Terms of Three Metrics: ACC, RMSE, and AUC Averaged on 10 Runs,  Obtained on The ASSISTments and SLP Datasets. The Best Results in Each Column  are Highlighted.}
	\setlength{\tabcolsep}{1.5mm}{
		\begin{tabular}{c|c|ccc|ccc|ccc|ccc}
			\toprule
			\multicolumn{2}{c}{\textbf{Train/test ratio}} & \multicolumn{3}{c}{\textbf{50\%/50\%}} & \multicolumn{3}{c}{\textbf{60\%/40\%}} & \multicolumn{3}{c}{\textbf{70\%/30\%}} & \multicolumn{3}{c|}{\textbf{80\%/20\%}}\\ % & \multicolumn{3}{c|}{\XG{\textbf{Friedman test   with Nemenyi post-hoc}}} \\
			\midrule
			\textbf{Dataset} & \textbf{Method} & \textbf{ACC} & \textbf{RMSE} & \multicolumn{1}{c}{\textbf{AUC}}& \textbf{ACC} & \textbf{RMSE} & \multicolumn{1}{c}{\textbf{AUC}}& \textbf{ACC} & \textbf{RMSE}& \multicolumn{1}{c}{\textbf{AUC}} & \textbf{ACC} & \textbf{RMSE} & \textbf{AUC} \\ % & \XG{\textbf{Method}}  & \XG{\textbf{Rank}} & \XG{\textbf{Symbol}} \\
			\midrule
			\ & \textbf{DINA} & 0.6532  & 0.4821  & 0.7103  & 0.6617  & 0.4771  & 0.7206  & 0.6672  & 0.4735  & 0.7260  & 0.6747  & 0.4699  & 0.7341 \\ %&  $\overline{\mathbf{A7}}$ & 2.17 &  \ding{178} \\
			& \textbf{IRT} & 0.6944  & 0.4595  & 0.7102  & 0.7006  & 0.4545  & 0.7203  & 0.7024  & 0.4536  & 0.7241  & 0.7058  & 0.4511  & 0.7285  \\ %&   $\overline{\mathbf{A6}}$ & 2.50 &  \ding{177}\\	 		
			& \textbf{MIRT} & 0.6885  & 0.5309  & 0.7141  & 0.6945  & 0.5285  & 0.7203  & 0.7069  & 0.5152  & 0.7412  & 0.7251  & 0.5005  & 0.7599 \\ %  & $\overline{\mathbf{A5}}$ & 4.50 &  \ding{176}\\	
			& \textbf{MF} & 0.6920  & 0.4577  & 0.7222  & 0.6927  & 0.4553  & 0.7242  & 0.6936  & 0.4542  & 0.7280  & 0.6930  & 0.4551  & 0.7244  \\ %& $\overline{\mathbf{A4}}$ & 3.17 &  \ding{175} \\	
			
			& \textbf{NCD} & 0.7155  & 0.4411  & 0.7396  & 0.7208  & 0.4380  & 0.7480  & 0.7243  & 0.4344  & 0.7548  & 0.7279  & 0.4321  & 0.7575  \\ %& $\overline{\mathbf{A3}}$ & 5.87 &  \ding{174} \\	
			& \textbf{RCD} & 0.7281  & {0.4255 } & 0.7618  & 0.7288  & 0.4244  & 0.7612  & 0.7307  & 0.4257  & 0.7655  & 0.7357  & 0.4210  & 0.7728  \\ %&   $\overline{\mathbf{A2}}$ & 6.62 & \ding{173} \\	
			
			& {\textbf{CDGK}} & {0.7121}  & {0.4398}   & {0.7360}  & {0.7192}  & {0.4344}  & {0.7454}   & {0.7226}   & {0.4338}   & {0.7506}   & {0.7342}   & {0.4251}   & {0.7567}  \\ % &   $\overline{\mathbf{A1}}$ & 9.17 & \ding{172} \\	
			& {\textbf{KSCD}} & {0.7167}   & {\hl{0.4245}}  & {0.7470}  & {0.7205} & {0.4239}  & {0.7534}  & {0.7249} &{0.4179}  & {0.7594}  & {0.7351}  & {0.4140}  & {0.7636} \\ % &   \textbf{DINA} & 14.17 &  $\mathcal{D}$ \\
			\cmidrule{2-14}          
			& \textbf{\textit{A1}}		 & 0.7101  & 0.4423  & 0.7410  & 0.7236  & 0.4367  & 0.7495  & 0.7246  & 0.4335  & 0.7528  & 0.7276  & 0.4301  & 0.7621 \\ % &  \multirow{2}[1]{*}{ \textbf{IRT} } &  \multirow{2}[1]{*}{ {13.13} }  &  \multirow{2}[1]{*}{  $\mathcal{I}$} \\ 
			& ($\overline{\mathbf{A1}}$) & 0.7126  & 0.4417 & 0.7414  & 0.7218  & 0.4369  & 0.7491  & 0.7260  & 0.4277  & 0.7549  & 0.7289  & 0.4319  & 0.7644 \\ % &   & &  \\	

			& \textbf{\textit{A2}} 		 & 0.7165  & 0.4336  & 0.7545  & 0.7238  & 0.4301  & 0.7610  & 0.7263  & 0.4287  & 0.7629  & 0.7452  & 0.4272  & 0.7726 \\ %  & \multirow{2}[1]{*}{ \textbf{MIRT} } &  \multirow{2}[1]{*}{ {13.38} }  &   \multirow{2}[1]{*}{ $\mathcal{M}$} \\ 
			%	& ($\overline{\mathbf{A2}}$) & 0.7150  & 0.4367  & 0.7533  & 0.7222  & 0.4311  & 0.7589  & 0.7285  & 0.4255  & 0.7651  & 0.7456  & 0.4266  & 0.7734  &    & &  \\	
			& ($\overline{\mathbf{A2}}$) & 0.7145 &0.4375 &0.7549 & 0.7726 & 0.4313 & 0.7593& 0.7276 & 0.4263 & 0.7655 &0.7449 & 0.4268 & 0.7731 \\ %&   & &  \\	

			\textbf{ASSIST}
			& \textbf{\textit{A3}} 		 & 0.7229  & 0.4328  & 0.7562  & 0.7305  & 0.4265  & 0.7647  & 0.7315  & 0.4263  & 0.7682  & 0.7353  & 0.4252  & 0.7766 \\ % &  \multirow{2}[1]{*}{ \textbf{MF} } &  \multirow{2}[1]{*}{ {13.17} }  &    \multirow{2}[1]{*}{ $\mathcal{F}$} \\ 			
			\textbf{ments}
			%	& ($\overline{\mathbf{A3}}$) & 0.7235  & 0.4305  & 0.7570  & 0.7317  & 0.4289  & 0.7660  & 0.7305  & 0.4276  & 0.7683  & 0.7355  & 0.4246  & 0.7758  &    & &  \\	
			& ($\overline{\mathbf{A3}}$) &  0.7237 & 0.4300 & 0.7567 & 0.7312 &  0.4285 & 0.7658 & 	0.7307 & 0.4281&  0.7688& 	0.7353&  0.4241 & 0.7763 \\ %&    & &  \\	

			& \textbf{\textit{A4}} 		 & 0.7415  & 0.4350  & 0.7602  & 0.7560  & 0.4238  & 0.7812  & 0.7674  & 0.4152  & 0.7965  & 0.7857  & 0.4001  & 0.8210 \\ % &  \multirow{2}[1]{*}{ \textbf{NCD} } &  \multirow{2}[1]{*}{ {10.17} }  &   \multirow{2}[1]{*}{ $\mathcal{N}$ }\\ 			
			%& ($\overline{\mathbf{A4}}$) & 0.7410  & 0.4343  & 0.7603  &  0.7567 & 0.4222  & 0.7806  & 0.7659  & 0.4170  & 0.7947  & 0.7860  & 0.3997  & 0.8207  & & &  \\	
			& ($\overline{\mathbf{A4}}$) & 0.7406  &0.4339  &0.7608 & 0.7561  &0.4220  &0.7809  &	0.7662 & 0.4173  &0.7945  &	0.7862  &0.3995 & 0.8211 \\ % & & &  \\	

			& \textbf{\textit{A5}} 		 & 0.7377  & 0.4397  & 0.7814  & 0.7530  & 0.4317  & 0.8019  & 0.7630  & 0.4215  & 0.8182  & 0.7810  & 0.4107  & 0.8409 \\ % &  \multirow{2}[1]{*}{ \textbf{RCD} } &  \multirow{2}[1]{*}{ {5.33} }  &  \multirow{2}[1]{*}{ $\mathcal{R}$} \\ 	
			%& ($\overline{\mathbf{A5}}$) & 0.7385  & 0.4402  & 0.7822  & 0.7533  & 0.4325  & 0.8011  & 0.7641  & 0.4200  & 0.8169  & 0.7815  & 0.4119  & 0.8413  &   & &  \\	
			
			& ($\overline{\mathbf{A5}}$) & 0.7388& 0.4409 &0.7825 &0.7536 &0.4331 &0.8014& 0.7634 &0.4206& 0.8174&	0.7817 &0.4123& 0.8414 \\ %&   & &  \\	

			& \textbf{\textit{A6}} 		 & 0.7441  & 0.4355  & \hl{0.7860 } & 0.7593  & 0.4252  & 0.8066  & 0.7712  & 0.4183  & 0.8244  & 0.7884  & 0.4018  & 0.8474 \\ %  & \multirow{2}[1]{*}{ \textbf{CDGK} } &  \multirow{2}[1]{*}{ {10.17} }  &  \multirow{2}[1]{*}{ $\mathcal{C}$}  \\ 	
			%	& ($\overline{\mathbf{A6}}$) & 0.7439  & 0.4360  &  0.7855      & 0.7582  & 0.4270  & 0.8049  & 0.7723  & 0.4201  & 0.8260  & 0.7879  & 0.4000  & 0.8479   &   & &  \\	
			
			& ($\overline{\mathbf{A6}}$) &	0.7442  & 0.4364  &0.7857  &	0.7585  &0.4264  &0.8045  &	0.7725 & 0.4207 & 0.8255  &	0.7880  &0.4003 & 0.8477 \\ %&   & &  \\	

			&\textbf{\textit{A7}} 		 & \hl{0.7438 } & 0.4372  & 0.7839  & \hl{0.7609 } & \hl{0.4237 } & \hl{0.8071 } & \hl{0.7735 } & \hl{0.4148 } & \hl{0.8250 } & \hl{0.7916 } & \hl{0.3993 } & \hl{0.8496 } \\ % & \multirow{2}[1]{*}{ \textbf{KSCD} } &  \multirow{2}[1]{*}{ {6.5} }  &   \multirow{2}[1]{*}{ $\mathcal{K}$ }\\ 	
			%	& ($\overline{\mathbf{A7}}$) & 0.7444       &  0.4366 & 0.7846  &  0.7615      & 0.4244       & 0.8075       & 0.7723       & 0.4135       & 0.8253       &  0.7915      & 0.3989       & 0.8500        &   & &  \\	
			& ($\overline{\mathbf{A7}}$) &	0.7441 & 0.4369 & 0.7847 &	0.7614 & 0.4248 & 0.8073 &	0.7727  & 0.4130 & 0.8250	& 0.7915  & 0.3996  & 0.8499 \\ %&   & &  \\	

			\midrule
			& 
			\textbf{DINA} & 0.6026  & 0.5200  & 0.6444  & 0.5735  & 0.5348  & 0.6240  & 0.6201  & 0.5063  & 0.6695  & 0.5871  & 0.5220  & 0.6565 \\ % &  $\overline{\mathbf{S6}}$ & 1.83 &  \\
			
			& \textbf{IRT} & 0.6048  & 0.4960  & 0.7061  & 0.6381  & 0.4779  & 0.7351  & 0.6502  & 0.4711  & 0.7340  & 0.6882  & 0.4540  & 0.7712 \\ % &  $\overline{\mathbf{S7}}$ & 2.25 &  \\
			& \textbf{MIRT} & 0.7145  & 0.4948  & 0.7595  & 0.7246  & 0.4842  & 0.7750  & 0.7376  & 0.4757  & 0.7852  & 0.7364  & 0.4772  & 0.7847 \\ % &   $\overline{\mathbf{S5}}$ & 3.25 &  \\
			& \textbf{MF} & 0.7655  & 0.4199  & 0.8229  & 0.7807  & 0.4081  & 0.8358  & 0.7743  & 0.4036  & 0.8372  & 0.7876  & 0.3997  & 0.8428  \\ % &  $\overline{\mathbf{S4}}$ & 2.92 &  \\
			
			& \textbf{NCD} & 0.7550  & 0.4425  & 0.8137  & 0.7741  & 0.4262  & 0.8330  & 0.7703  & 0.4246  & 0.8326  & 0.7780  & 0.4012  & 0.8397 \\ % &  $\overline{\mathbf{S3}}$ & 6.25 &  \\
			
			& 	{\textbf{CDGK}} & {0.7544}  & {0.4429} & {0.8130}  & {0.7755}  & {0.4244}  & {0.8334}  & {0.7716}  & {0.4257} & {0.8313}  & {0.7879}  & {0.3844}  &{0.8421} \\ % &   $\overline{\mathbf{S2}}$ & 5.42 &  \\	
			& {\textbf{KSCD}} & {0.7563} & {0.4258} & {0.8218}  & {0.7738}  & {0.4125}  & {0.8391} & {0.7710}  & {0.4085}  & {0.8377}  & {0.7857} & {0.3844} & {0.8464}  \\ %&  $\overline{\mathbf{S1}}$ & 8.67 &  \\	
			
			\cmidrule{2-14}      
			
			& \textbf{\textit{S1}} 		 & 0.7674  & 0.4279  & 0.8224  & 0.7746  & 0.4183  & 0.8321  & 0.7756  & 0.4210  & 0.8345  & 0.7872  & 0.4093  & 0.8439  \\ %& \multirow{2}[1]{*}{ \textbf{DINA} } &  \multirow{2}[1]{*}{ {14.00} }  &  \multirow{2}[1]{*}{ null} \\ 
			%& ($\overline{\mathbf{S1}}$) & 0.7678  & 0.4285  & 0.8229  & 0.7733  & 0.4170  & 0.8315  & 0.7740  & 0.4222  & 0.8339  & 0.7885  & 0.4079  & 0.8449  &  & &  \\	
			& ($\overline{\mathbf{S1}}$) &  0.7677 &   0.4286 & 0.8230& 	0.7739&  0.4166 &  0.8320 & 	0.7737 &  0.4214 &  0.8336 & 	0.7883 &  0.4085 &  0.8446 \\ % &  & &  \\	

			& \textbf{\textit{S2}} 		 & 0.7741  & 0.4219  & 0.8246  & 0.7840  & 0.4051  & 0.8441  & 0.7856  & 0.3942  & 0.8497  & 0.7915  & 0.3912  & 0.8560 \\ % &  \multirow{2}[1]{*}{ \textbf{IRT} } &  \multirow{2}[1]{*}{ {12.75} }  &  \multirow{2}[1]{*}{ null} \\  
			%& ($\overline{\mathbf{S2}}$) & 0.7733  & 0.4213  & 0.8239  & 0.7842  & 0.4039  & 0.8450  & 0.7866  & 0.3943  & 0.8511  & 0.7917  & 0.3920  & 0.8565  &  & &  \\	
			& ($\overline{\mathbf{S2}}$) &  0.7731 & 0.4218 & 0.8237 &	0.7843 & 0.4034 & 0.8448  &	0.7868 & 0.3939& 0.8510 &	0.7921& 0.3922 & 0.8563 \\ %&  & &  \\	

			\textbf{SLP}	& \textbf{\textit{S3}} 		 & 0.7722  & 0.4215  & 0.8349  & 0.7735  & 0.4110  & 0.8457  & 0.7772  & 0.3997  & 0.8534  & 0.7855  & 0.3912  & 0.8572  \\ % & \multirow{2}[1]{*}{ \textbf{MIRT} } &  \multirow{2}[1]{*}{ {12.25} }  &  \multirow{2}[1]{*}{ null} \\ 
			%	& ($\overline{\mathbf{S3}}$) & 0.7730  & 0.4203  & 0.8355  & 0.7740  & 0.4101  & 0.8445  & 0.7765  & 0.3993  & 0.8527  & 0.7853  & 0.3919  & 0.8579  &    & &  \\		
			& ($\overline{\mathbf{S3}}$) & 0.7733 &0.4206 &0.8359 &	0.7743 &0.4112& 0.8444 &	0.7769& 0.4000 &0.8524 &	0.7855& 0.3913& 0.8577 \\ %&    & &  \\

			& \textbf{\textit{S4}} 		 & \hl{0.7806 } & \hl{0.3933 } & 0.8429  & 0.7916  & 0.3871  & 0.8539  & 0.7861  & 0.3904  & 0.8554  & 0.7964  & 0.3797  & 0.8635 \\ % & \multirow{2}[1]{*}{ \textbf{MF} } &  \multirow{2}[1]{*}{ {7.42} }  &  \multirow{2}[1]{*}{ null} \\ 		
			%& ($\overline{\mathbf{S4}}$) & 0.7810       & 0.3922       &  0.8437 & 0.7900  & 0.3897  & 0.8528  & 0.7880  & 0.3887  & 0.8570  & 0.7969  & 0.3790  & 0.8633  & & &  \\	
			& ($\overline{\mathbf{S4}}$) &  		0.7812 & 0.3928 & 0.8441		& 	0.7900 & 0.3891 & 0.8522		& 	0.7883 & 0.3895 & 0.8566		& 	0.7966 & 0.3799 & 0.8637 \\ % & & &  \\	

			& \textbf{\textit{S5}} 		 & 0.7789  & 0.3934  & 0.8439  & 0.7933  & \hl{0.3840 } & 0.8561  & 0.7832  & 0.3863  & 0.8559  & 0.7962  & 0.3794  & 0.8641 \\ % & \multirow{2}[1]{*}{ \textbf{NCD} } &  \multirow{2}[1]{*}{ {10.25} }  &  \multirow{2}[1]{*}{ null} \\ 	 
			%& ($\overline{\mathbf{S5}}$) & 0.7780  & 0.3960  & 0.8428  & 0.7924  & 0.3861       & 0.8553  &  0.7840 & 0.3871  & 0.8570  & 0.7955  & 0.3802  & 0.8647 & & &  \\		
			& ($\overline{\mathbf{S5}}$) & 0.7786 & 0.3949 & 0.8433 & 0.7930  & 0.3847& 0.8566 & 0.7831  &0.3860 & 0.8555 &	0.7949 & 0.3813  & 0.8642 \\ %& & &  \\	

			& \textbf{\textit{S6}} 		 & 0.7802  & 0.3930  & \hl{0.8470 } & \hl{0.7955} & 0.3844  & 0.8579  & 0.7828  & 0.3877  & 0.8580  & 0.8001  & \hl{0.3773 } & 0.8669  \\ %& \multirow{2}[1]{*}{ \textbf{CDGK} } &  \multirow{2}[1]{*}{ {9.58} }  &  \multirow{2}[1]{*}{ null} \\ 	 			
			%	& ($\overline{\mathbf{S6}}$) & 0.7810  & 0.3919  &  0.8475      & 0.7960      & 0.3858  & 0.8582  & 0.7812  & 0.3899  & 0.8555  & 0.7989  &  0.3788      & 0.8660  & & &  \\
			& ($\overline{\mathbf{S6}}$) & 0.7803& 0.3927& 0.8475 &0.7953 &0.3844 &0.8581 	&0.7810 &0.3888 &0.8567 	&0.7987 &0.3779 &0.8663\\ % & & &  \\

			& \textbf{\textit{S7}} 		 & 0.7786  & 0.3973  & 0.8446  & 0.7941  & 0.3865  & \hl{0.8587} & \hl{0.7854 } & \hl{0.3833} & \hl{0.8621} & \hl{0.8023 } & 0.3781  & \hl{0.8685}  \\ %& \multirow{2}[1]{*}{ \textbf{KSCD} } &  \multirow{2}[1]{*}{ {8.17} }  &  \multirow{2}[1]{*}{ null} \\ 	
			%& ($\overline{\mathbf{S7}}$) & 0.7777  & 0.3999  & 0.8435  & 0.7945  & 0.3870  & 0.8579      &  0.7839      & 0.3840      &  0.8615     &   0.8025     & 0.3789  & 0.8677  &  & &  \\
			& ($\overline{\mathbf{S7}}$) &  0.7779 & 0.4002 & 0.8435  & 	0.7945 & 0.3866&  0.8577   & 	0.7841&  0.3844&  0.8622  & 	0.8031 & 0.3796 &  0.8683 \\ %&  & &  \\
			\bottomrule
	\end{tabular}			}
	\label{tab:overall result}%
\end{table*}%

\subsubsection{\textbf{Datasets}}\indent
To verify the effectiveness of the proposed EMO-NAS-CD, we conducted  experiments on two real-world education datasets, including ASSISTments~\cite{feng2009addressing} and SLP~\cite{slp2021edudata}.
We have summarized the statistics of two datasets in Table~\ref{tab:dataset} and presented the descriptions of two datasets as follows:
\begin{itemize}
	\footnotesize
	\item \textbf{ASSISTments} (\textit{ASSISTments 2009-2010 skill builder})~\cite{feng2009addressing} is an openly available dataset created in	2009 by the ASSISTments online tutoring service system. 
	Here we adopted the public corrected version that does not contain  duplicate data.
	As can be seen, there are  more than 4 thousand students, nearly 18 thousand exercises, and  over 300 thousand response logs in the dataset.
	%~\cite{xiong2016going}
	
	\item  {\textbf{SLP} (\textit{Smart Learning Partner})}~\cite{slp2021edudata} is another  public education dataset  published	in 2021.
	SLP collects the regularly captured academic performance data of learners during their three-year study on eight different subjects, including Chinese, mathematics, English, physics, chemistry, biology, history and geography.
	The dataset contains nearly 58 thousand response logs of 1,499 students on 907 exercises.	
\end{itemize}

According to the experiences of previous work~\cite{wang2020neural,gao2021rcd,wang2021using}, 
we filtered out students with less than 15 response logs for all datasets to ensure that 
there are sufficient data to model each student  for diagnosis.

\begin{table*}[htbp]
	\centering
	\caption{Friedman test with Nemenyi post-hoc procedure on the results of  Table~\ref{tab:overall result}.
		Architectures $\overline{{A1}}$  to $\overline{{A7}}$ are denoted by \ding{172} to \ding{178},  
		$\overline{{S1}}$  to $\overline{{S7}}$ are denoted by \ding{182} to \ding{188}, 
		DINA, IRT, MIRT, MF, NCD, RCD, CDGK, and KSCD are denoted by   $\mathcal{D}$, $\mathcal{I}$, $\mathcal{M}$ , $\mathcal{F}$     , $\mathcal{N}$, $\mathcal{R}$, 
		$\mathcal{C}$, and $\mathcal{K}$, respectively. 
		Here left and right tables summarize the  analysis results on the ASSISTments and SLP, respectively, 
		where  ‘1’ indicates significant difference and '0' otherwise. }
	
			\renewcommand{\arraystretch}{0.9}
	\setlength{\tabcolsep}{0.9mm}{
	\begin{tabular}{c|ccccccccccccccc|c}
			\hline
			\multicolumn{17}{c}{\textbf{Friedman Test with Nemenyi Post-hoc on the ASSISTments}}   \\
		\hline
		
		   \textbf{Method} & \ding{178}     & \ding{177}     & \ding{175}     & \ding{176}     &  $\mathcal{R}$   & \ding{174}      & $\mathcal{K}$     & \ding{172}      & \ding{172}         & $\mathcal{N}$      & $\mathcal{C}$      & $\mathcal{I}$      & $\mathcal{F}$     & $\mathcal{M}$     & $\mathcal{D}$  & \textbf{Rank} \\
	\hline    
		\ding{178}      & -     & 0     & 0     & 0     & 0     & 0     & 0     & 0     & 1     & 1     & 1     & 1     & 1     & 1     & 1     & 2.08  \\
		\ding{177}      & 0     & -     & 0     & 0     & 0     & 0     & 0     & 0     & 1     & 1     & 1     & 1     & 1     & 1     & 1     & 2.67  \\
		\ding{175}      & 0     & 0     & -     & 0     & 0     & 0     & 0     & 0     & 0     & 1     & 1     & 1     & 1     & 1     & 1     & 3.17  \\
		\ding{176}      & 0     & 0     & 0     & -     & 0     & 0     & 0     & 0     & 0     & 0     & 0     & 1     & 1     & 1     & 1     & 4.50  \\
		 $\mathcal{R}$      & 0     & 0     & 0     & 0     & -     & 0     & 0     & 0     & 0     & 0     & 0     & 1     & 1     & 1     & 1     & 5.33  \\
		\ding{174}       & 0     & 0     & 0     & 0     & 0     & -     & 0     & 0     & 0     & 0     & 0     & 1     & 1     & 1     & 1     & 5.88  \\
		$\mathcal{K}$      & 0     & 0     & 0     & 0     & 0     & 0     & -     & 0     & 0     & 0     & 0     & 1     & 1     & 1     & 1     & 6.50  \\
		\ding{173}     & 0     & 0     & 0     & 0     & 0     & 0     & 0     & -     & 0     & 0     & 0     & 1     & 1     & 1     & 1     & 6.54  \\
		\ding{172}     & 1     & 1     & 0     & 0     & 0     & 0     & 0     & 0     & -     & 0     & 0     & 0     & 0     & 0     & 0     & 9.17  \\
		$\mathcal{N}$     & 1     & 1     & 1     & 0     & 0     & 0     & 0     & 0     & 0     & -     & 0     & 0     & 0     & 0     & 0     & 10.17  \\
		$\mathcal{C}$     & 1     & 1     & 1     & 0     & 0     & 0     & 0     & 0     & 0     & 0     & -     & 0     & 0     & 0     & 0     & 10.17  \\
		$\mathcal{I}$     & 1     & 1     & 1     & 1     & 1     & 1     & 1     & 1     & 0     & 0     & 0     & -     & 0     & 0     & 0     & 13.13  \\
		$\mathcal{F}$     & 1     & 1     & 1     & 1     & 1     & 1     & 1     & 1     & 0     & 0     & 0     & 0     & -     & 0     & 0     & 13.17  \\
		$\mathcal{M}$     & 1     & 1     & 1     & 1     & 1     & 1     & 1     & 1     & 0     & 0     & 0     & 0     & 0     & -     & 0     & 13.38  \\
		$\mathcal{D}$     & 1     & 1     & 1     & 1     & 1     & 1     & 1     & 1     & 0     & 0     & 0     & 0     & 0     & 0     & -     & 14.17  \\
		\hline
		\multicolumn{4}{c|}{$p$-values}  & \multicolumn{5}{c}{2.29e-24}          & \multicolumn{8}{|c}{Significance level $\alpha$ 0.05}  \\
		\bottomrule
	\end{tabular}%
}	
\setlength{\tabcolsep}{1.1mm}{
	\begin{tabular}{c|cccccccccccccc|c}
		\hline
		\multicolumn{16}{c}{\textbf{Friedman Test with Nemenyi Post-hoc on the SLP}}   \\
		\hline
		\textbf{Method} & \ding{187}    & \ding{188}      & \ding{185}      & \ding{186}      & \ding{183}      & \ding{184}      & $\mathcal{F}$     & $\mathcal{K}$     & \ding{182}     & $\mathcal{C}$    & $\mathcal{N}$    & $\mathcal{M}$    & $\mathcal{I}$    & $\mathcal{D}$    & \textbf{Rank} \\
		\hline
		\ding{187}      & -     & 0     & 0     & 0     & 0     & 0     & 0     & 1     & 1     & 1     & 1     & 1     & 1     & 1     & 1.83  \\
		\ding{188}      & 0     & -     & 0     & 0     & 0     & 0     & 0     & 1     & 1     & 1     & 1     & 1     & 1     & 1     & 2.25  \\
		\ding{185}      & 0     & 0     & -     & 0     & 0     & 0     & 0     & 0     & 1     & 1     & 1     & 1     & 1     & 1     & 2.92  \\
		\ding{186}      & 0     & 0     & 0     & -     & 0     & 0     & 0     & 0     & 0     & 1     & 1     & 1     & 1     & 1     & 3.25  \\
		\ding{183}      & 0     & 0     & 0     & 0     & -     & 0     & 0     & 0     & 0     & 0     & 0     & 1     & 1     & 1     & 5.42  \\
		\ding{184}      & 0     & 0     & 0     & 0     & 0     & -     & 0     & 0     & 0     & 0     & 0     & 1     & 1     & 1     & 6.25  \\
		$\mathcal{F}$      & 0     & 0     & 0     & 0     & 0     & 0     & -     & 0     & 0     & 0     & 0     & 0     & 0     & 1     & 7.42  \\
		$\mathcal{K}$      & 1     & 1     & 0     & 0     & 0     & 0     & 0     & -     & 0     & 0     & 0     & 0     & 0     & 1     & 8.17  \\
		\ding{182}     & 1     & 1     & 1     & 0     & 0     & 0     & 0     & 0     & -     & 0     & 0     & 0     & 0     & 0     & 8.67  \\
		$\mathcal{C}$    & 1     & 1     & 1     & 1     & 0     & 0     & 0     & 0     & 0     & -     & 0     & 0     & 0     & 0     & 9.58 \\
		$\mathcal{N}$    & 1     & 1     & 1     & 1     & 0     & 0     & 0     & 0     & 0     & 0     & -     & 0     & 0     & 0     & 10.25  \\
		$\mathcal{M}$    & 1     & 1     & 1     & 1     & 1     & 1     & 0     & 0     & 0     & 0     & 0     & -     & 0     & 0     & 12.25  \\
		$\mathcal{I}$    & 1     & 1     & 1     & 1     & 1     & 1     & 0     & 0     & 0     & 0     & 0     & 0     & -     & 0     & 12.75  \\
		$\mathcal{D}$    & 1     & 1     & 1     & 1     & 1     & 1     & 1     & 1     & 0     & 0     & 0     & 0     & 0     & -     & 14.00  \\
		\hline
			\multicolumn{4}{c|}{$p$-values}  & \multicolumn{5}{c}{2.62e-24}          & \multicolumn{7}{|c}{Significance level $\alpha$ 0.05}  \\
		\bottomrule
	\end{tabular}%
}

	\label{tab:addlabel}%
\end{table*}%

\subsubsection{\textbf{Compared Approaches and Metrics}}\indent
To validate the effectiveness of the proposed approach,
we compared  the diagnostic function architectures found by the proposed EMO-NAS-CD with state-of-the-art CDMs, 
 including   DINA~\cite{Torre2009DINA}, IRT~\cite{embretson2013item}, and MIRT~\cite{reckase2009multidimensional}, MF~\cite{koren2009matrix}, NCD~\cite{wang2020neural},  RCD~\cite{gao2021rcd}, CDGK~\cite{wang2021using}, and KSCD~\cite{ma2022knowledge}.
 % ECD? IRR?
 The detailed descriptions of these comparison CDMs can be found in Section~\ref{sec:related CD}.
 The source codes of most compared approaches are available at \url{https://github.com/orgs/bigdata-ustc/repositories}.
 {Note that 
the results of RCD on SLP are not reported since RCD needs extra manually enhanced inputs that SLP does not have.}
 
To measure the performance obtained by all CDMs, three evaluation metrics including  AUC, \textit{accuracy} (ACC), and \textit{root mean square error} (RMSE) are adopted.% ~\cite{chai2014root}
\subsubsection{\textbf{Parameter Settings}}\indent

\iffalse
\begin{figure*}[t]
	\centering
	\subfloat[Boxplot on the ASSISTments.]{\includegraphics[width=0.5\linewidth]{ASSIST.pdf}}
	\subfloat[Boxplot on the SLP.]{\includegraphics[width=0.5\linewidth]{SLP.pdf}}
	\caption{\XG{Boxplots for AUC values (under the  ratio of 80\%/20\%)  of comparison CDMs and seven found architectures on both datasets.}}
	\label{fig:boxplot}
\end{figure*}
\fi

\begin{itemize}
	\footnotesize
	\item \textbf{{1. Architecture Settings}} The dimension $D$ is equal to the number of knowledge concepts $K$, 
	$H_1$, $H_2$, and $H_3$ are set to 512, 256, and 1, respectively.
	
	\item \textbf{{2. Search Settings}} During the search process in the proposed EMO-NAS-CD, 
	each student's  response logs in each dataset  are randomly split into 70\%, 10\%, and 20\% as training, validating, and testing datasets, respectively. 
	To train the architecture encoded by each individual, 
	the \textit{Adam} optimizer with a learning rate  of 0.001 is used to optimize the Cross-Entropy loss between the prediction results and the targets, 
	where the size of each batch is set to 128, and the number of training epochs $Num_{E}$  is set to 30.
	For the proposed EMO-NAS-CD, the population size $Pop$ is set to 100, the maximal number of generation $Gen$ is set to 100, and the initial node range $Node_{range}$ is set to $\{2,4\}$.  
	
\item	\textbf{{3. Training Settings}} %To demonstrate the effectiveness of the found architectures on the different extents of data sparsity,
	For more convincing results, 
	we adopted multiple different settings  to split the dataset into training and test datasets for evaluating the model performance, 
	where the settings contain  50\%/50\%, 60\%/40\%, 70\%/30\%, and 80\%/20\% as suggested in~\cite{gao2021rcd}.
	Each found architecture needs to be retrained from scratch for 50 epochs, 
	the settings are the same as that in the above search settings.
	For a fair comparison, the parameter settings of all comparison CDMs are the same as those in their original papers to hold their best performance.

\end{itemize}

%All approaches and models are implemented with \textit{Pytorch} under Python, 
All experiments were conducted on a NVIDIA RTX 3090 GPU. 
Our source code can be  available at \url{https://github.com/DevilYangS/EMO-NAS-CD}.

\iffalse
\begin{figure*}
	\centering
	\subfloat{\includegraphics[width=0.2640\linewidth]{pareto_ASSIST_.pdf}}%0.43
	\subfloat{\includegraphics[width= 0.3760\linewidth]{Architecture_ASSIST_.pdf}}	
	\caption{\textbf{Left}: The non-dominated individuals searched by the proposed EMO-NAS-CD on the ASSISTments dataset. \textbf{Right}: the architecture visualization of architecture $A1$ to architecture $A7$.% which  correspond to the architectures from EMO-NAS-CD-A1 to EMO-NAS-CD-A7, respectively.
	}
	\label{fig:paretoASSIST}
\end{figure*}

\begin{figure*}
	\centering
	\subfloat{\includegraphics[width=0.2856\linewidth]{pareto_SLP_.pdf}}
	\subfloat{\includegraphics[width=0.36\linewidth]{Architecture_SLP_.pdf}}	
	\caption{
		\textbf{Left}: The non-dominated individuals searched by the proposed EMO-NAS-CD on the SLP dataset. \textbf{Right}: the architecture visualization of architecture $S1$ to architecture $S7$.		% which  correspond to the architectures from EMO-NAS-CD-S1 to EMO-NAS-CD-S7, respectively.	
	}
	\label{fig:paretoslp}
\end{figure*}

\fi

\subsection{Effectiveness of The Proposed EMO-NAS-CD}\label{sec:Eff}
%To verify the effectiveness of the diagnostic function architectures found by the proposed EMO-NAS-CD,
Table~\ref{tab:overall result} summarizes  the prediction performance comparison between the proposed EMO-NAS-CD and comparison CDMs in terms of ACC, RMSE, and AUC values that are averaged on 30 independent runs on the two datasets, where  five different splitting settings are considered.
Here   seven architectures (with different degrees of model interpretability) found by  EMO-NAS-CD in a single run are selected for comparison, 
where  architectures \textit{A1} to \textit{A7} are found on the ASSISTments and architectures \textit{S1} to  \textit{S7} are found on the SLP.
To this end, for some architectures that have similar model interpretability, 
the architecture with the best performance among these architectures will be selected for final comparison. 
For more convincing explanations,  
Table~\ref{tab:overall result} further shows the results of $\overline{{A1}}$ ($\overline{{S1}}$) to $\overline{{A7}}$ ($\overline{{S7}}$). 
$\overline{{A1}}$ refers to the  average results on ten different runs of EMO-NAS-CD. 
In each run, the architecture that has  similar interpretability  to \textit{A1}   is used to compute  $\overline{{A1}}$, which is the same to obtain $\overline{{A2}}$ to $\overline{{A7}}$ and   $\overline{{S1}}$ to $\overline{{S7}}$.
Besides, the Friedman test with Nemenyi procedure~\cite{derrac2011practical} (under significance level $\alpha$=0.05) was conducted on the results of comparison CDMs and $\overline{{A1}}$ ($\overline{{S1}}$) to $\overline{{A7}}$ ($\overline{{S7}}$), which is a nonparametric statistical procedure to check whether a set of samples are statistically different.
Table~\ref{tab:addlabel} summarizes the   statistical  results including significance analysis and rank of each method, where '1' indicates significant difference between two methods and '0' otherwise.

As can be observed from Table III and Table IV, 
nearly all architectures found by  EMO-NAS-CD (except for the simplest architectures \textit{A1} and \textit{S1}) 
exhibit significantly better performance than all comparison CDMs. 
Take the results under the splitting setting of 80\%/20\% for analysis, 
and the boxplots for AUC values (under this setting) of  comparison CDMs and seven found architectures  are further presented in Fig.\ref{fig:boxplot} for explicit observation.
As can be seen,
the most effective architecture  \textit{A7} outperforms the current best CDM (RCD) by over 0.07 on the ASSISTments dataset in terms of the AUC value.
Even for the simplest architecture \textit{A1}, 
there still holds the superiority of performance over most CDMs, 
which is competitive to KSCD and only  worse than  RCD, but KSCD and RCD use extra input information to enhance the performance.
Therefore, compared to the CDMs that do not have such input information, 
the performance difference between  our best-found architectures and these CDMs is more significant: 
the performance leading of \textit{A7} over  the best of these CDMs is up to 0.08 in terms of  AUC values, and  architecture \textit{A1} also outperforms  these CDMs.
It can be seen that the proposed approach achieves such a tremendous performance improvement by only designing more effective architectures  without extra input information.
In addition, we can find that the standard deviation of the proposed approach is very small from the comparisons 
between \textit{A1}  to \textit{A7}   and $\overline{{A1}}$ to $\overline{{A7}}$.
We can make the same observations and  conclusions based on the  results on the SLP dataset.

\iffalse
\begin{figure*}[t]
	\centering
	\subfloat[Boxplot on the ASSISTments.]{\includegraphics[width=0.57\linewidth]{ASSIST+SLP_1.pdf}}
	\subfloat[Boxplot on the SLP (excluding DINA, IRT, and MIRT due to their poor performance).]{\includegraphics[width=0.4\linewidth]{ASSIST+SLP_2.pdf}}
	\caption{\XG{Boxplots for AUC values (under the  ratio of 80\%/20\%)  of comparison CDMs and seven found architectures on both datasets.}}
	\label{fig:boxplot}
\end{figure*}
\fi

\begin{figure}[t]
	\centering
	\subfloat[Boxplot on the ASSISTments.]{\includegraphics[width=1.0\linewidth]{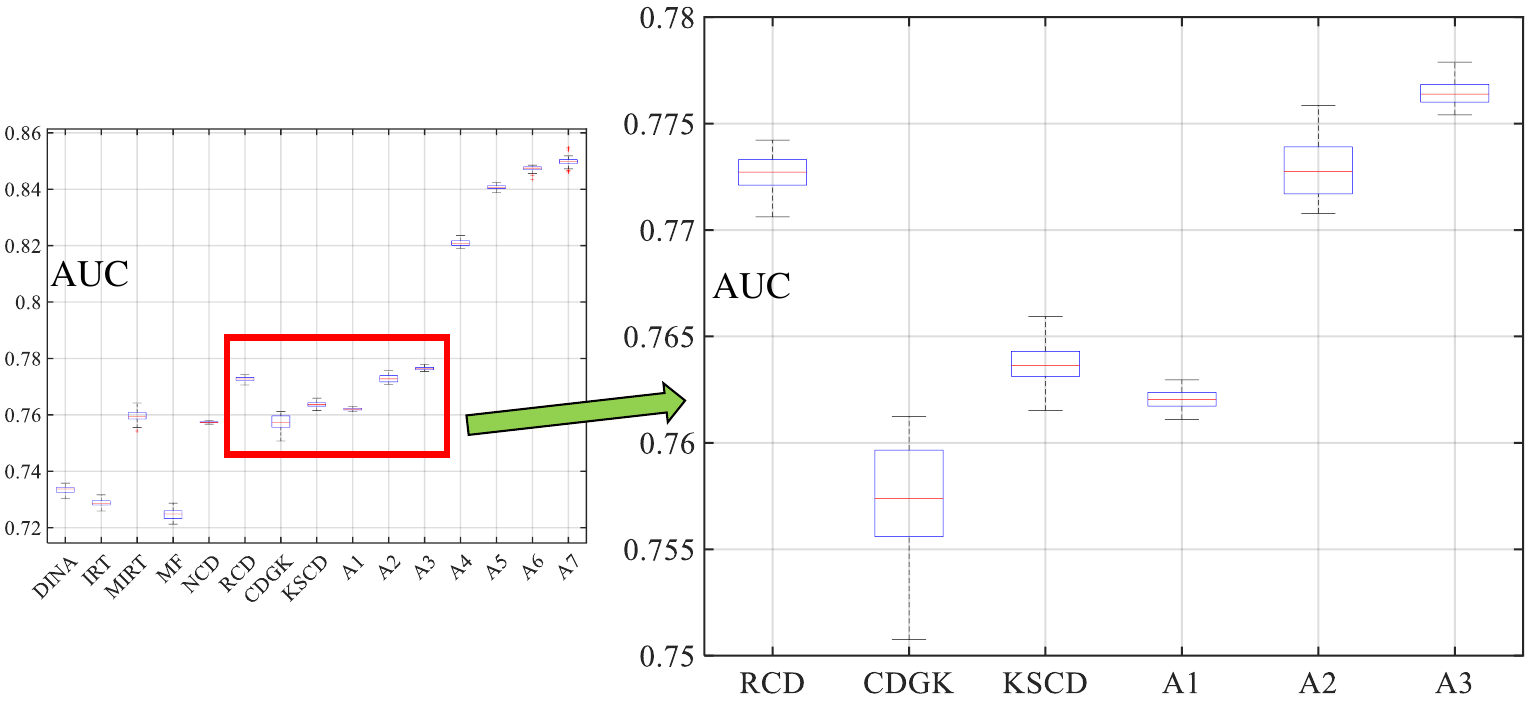}}\\
	\subfloat[Boxplot on the SLP (excluding DINA, IRT, and MIRT due to their poor performance).]{\includegraphics[width=0.7\linewidth]{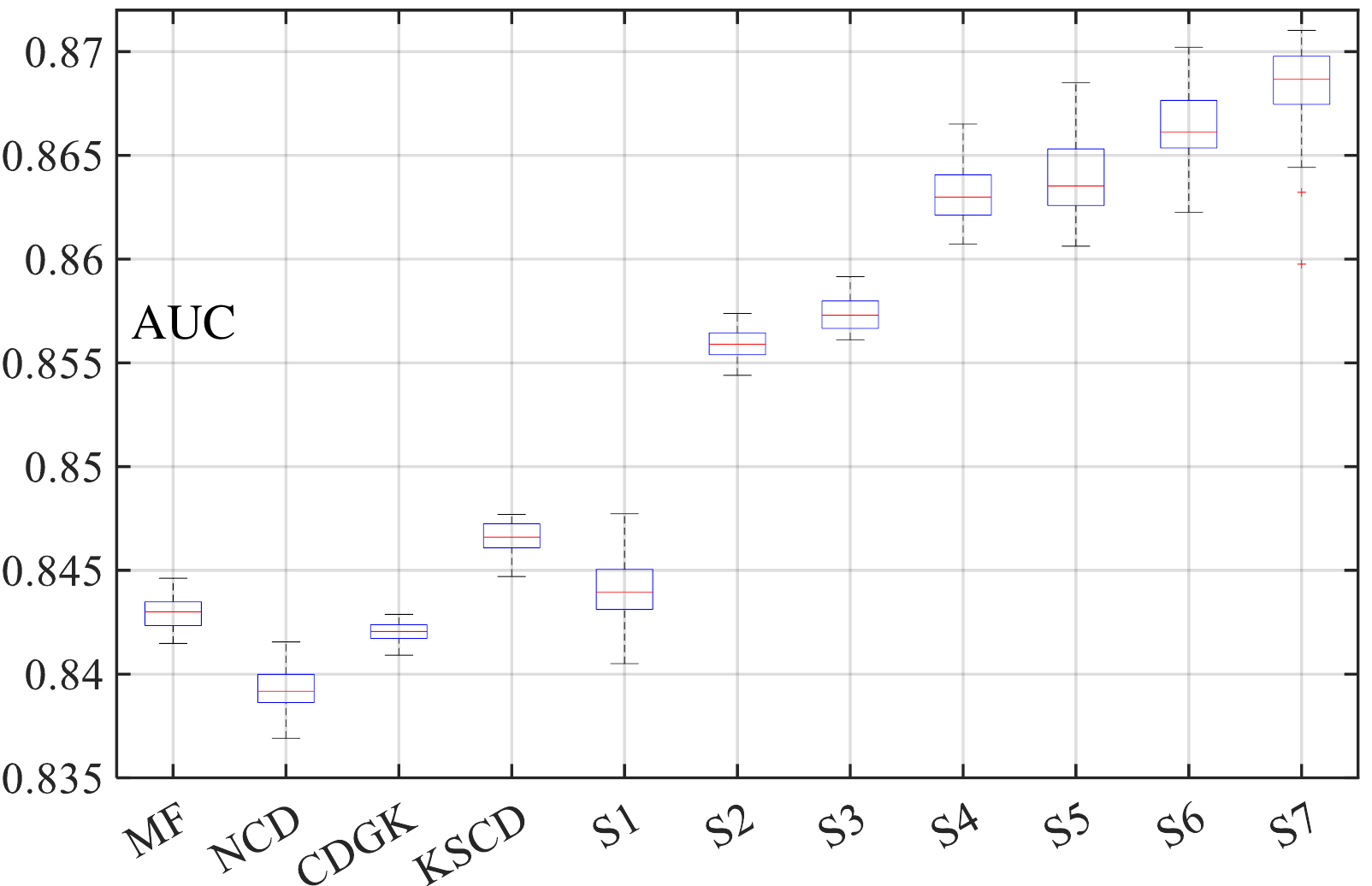}}
	\caption{Boxplots for AUC values (under the  ratio of 80\%/20\%)  of comparison CDMs and seven found architectures on both datasets.}
	\label{fig:boxplot}
\end{figure}

\begin{figure}
	\centering
	\subfloat{\includegraphics[width=0.4125\linewidth]{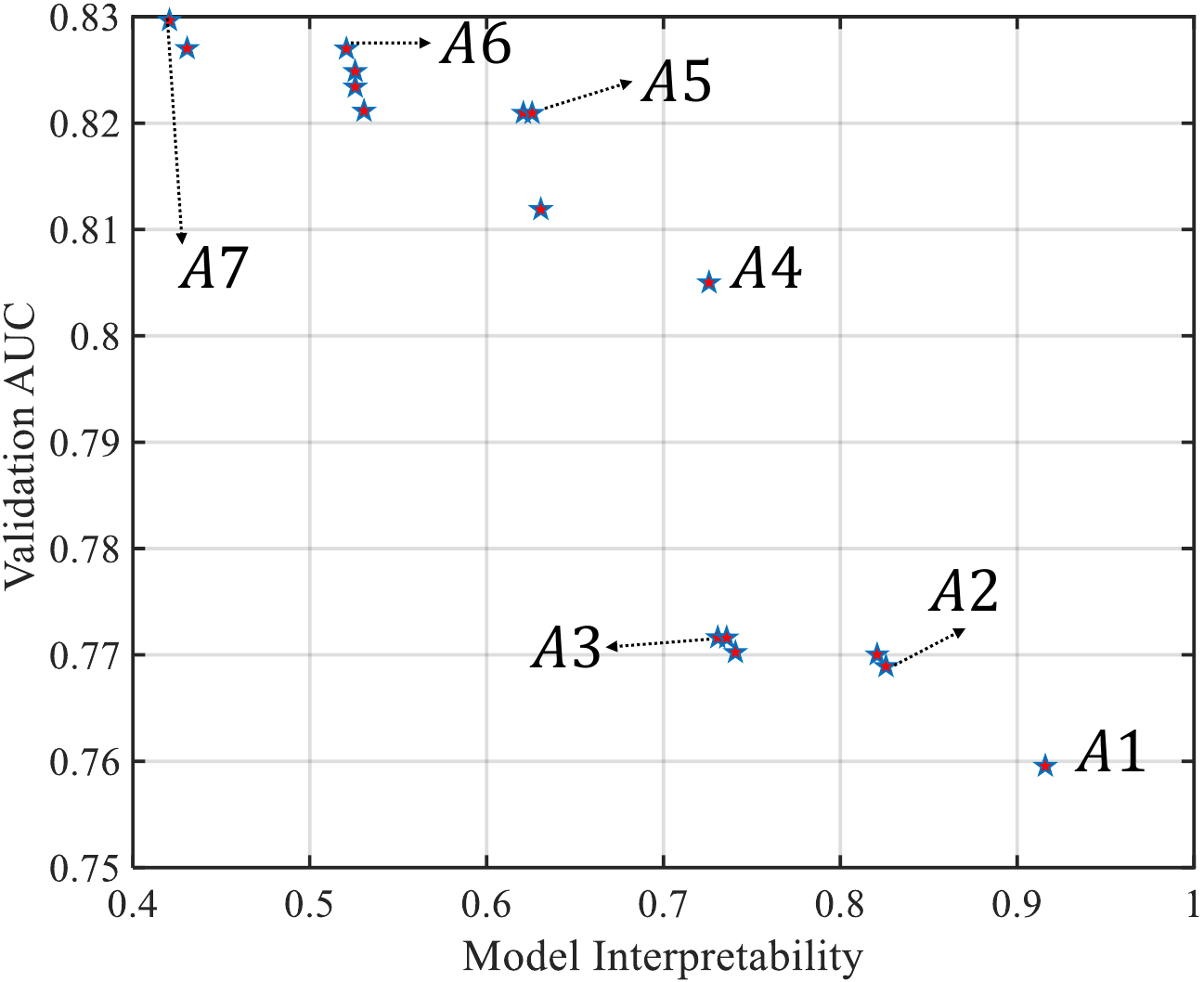}}%0.43
	\subfloat{\includegraphics[width= 0.5870\linewidth]{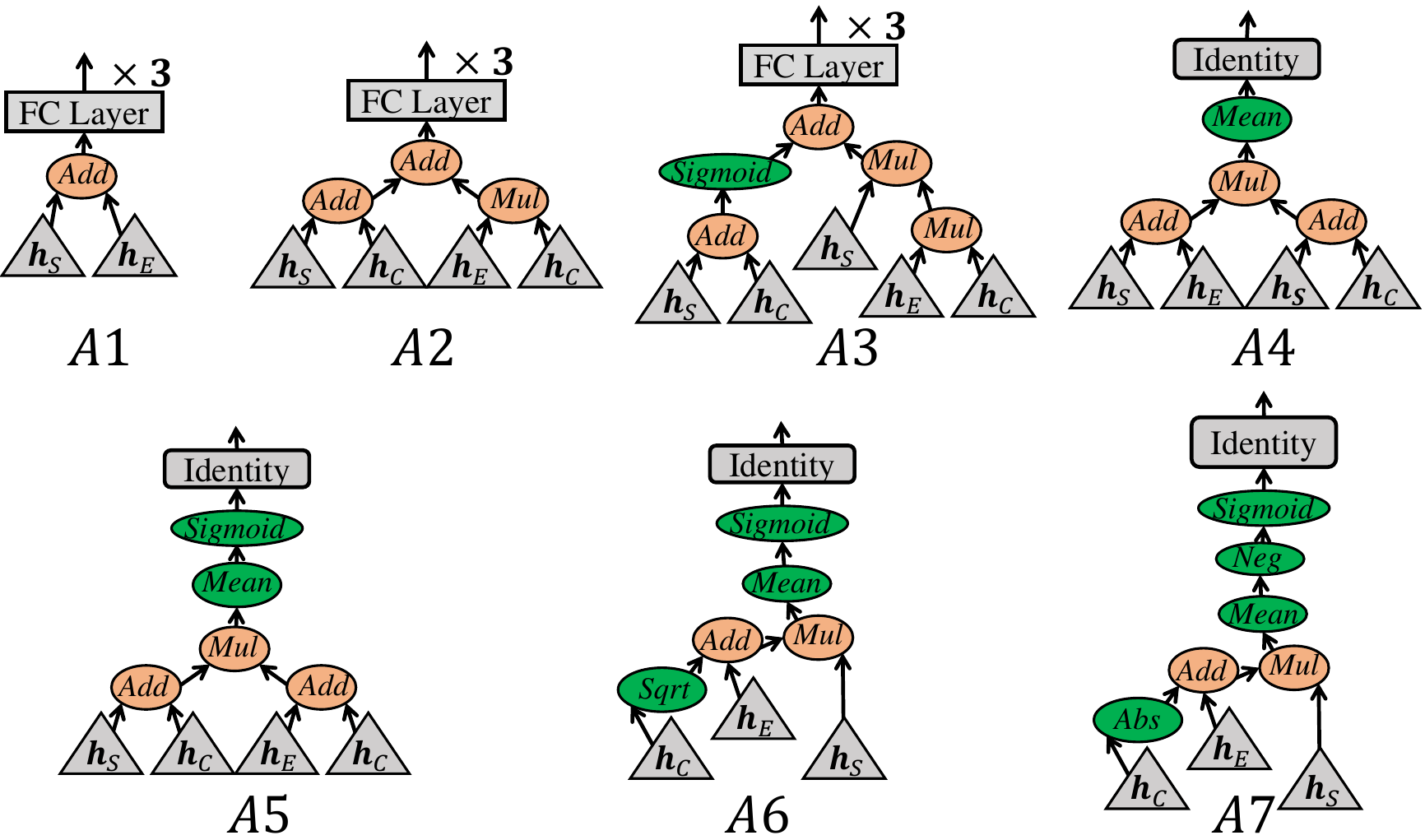}}	
	\caption{The non-dominated individuals found by the proposed EMO-NAS-CD on the ASSISTments  and  the  visualization of architecture $A1$ to architecture $A7$.% which  correspond to the architectures from EMO-NAS-CD-A1 to EMO-NAS-CD-A7, respectively.
	}
	\label{fig:paretoASSIST}
\end{figure}

\begin{figure}
	\centering
	\subfloat{\includegraphics[width=0.4125\linewidth]{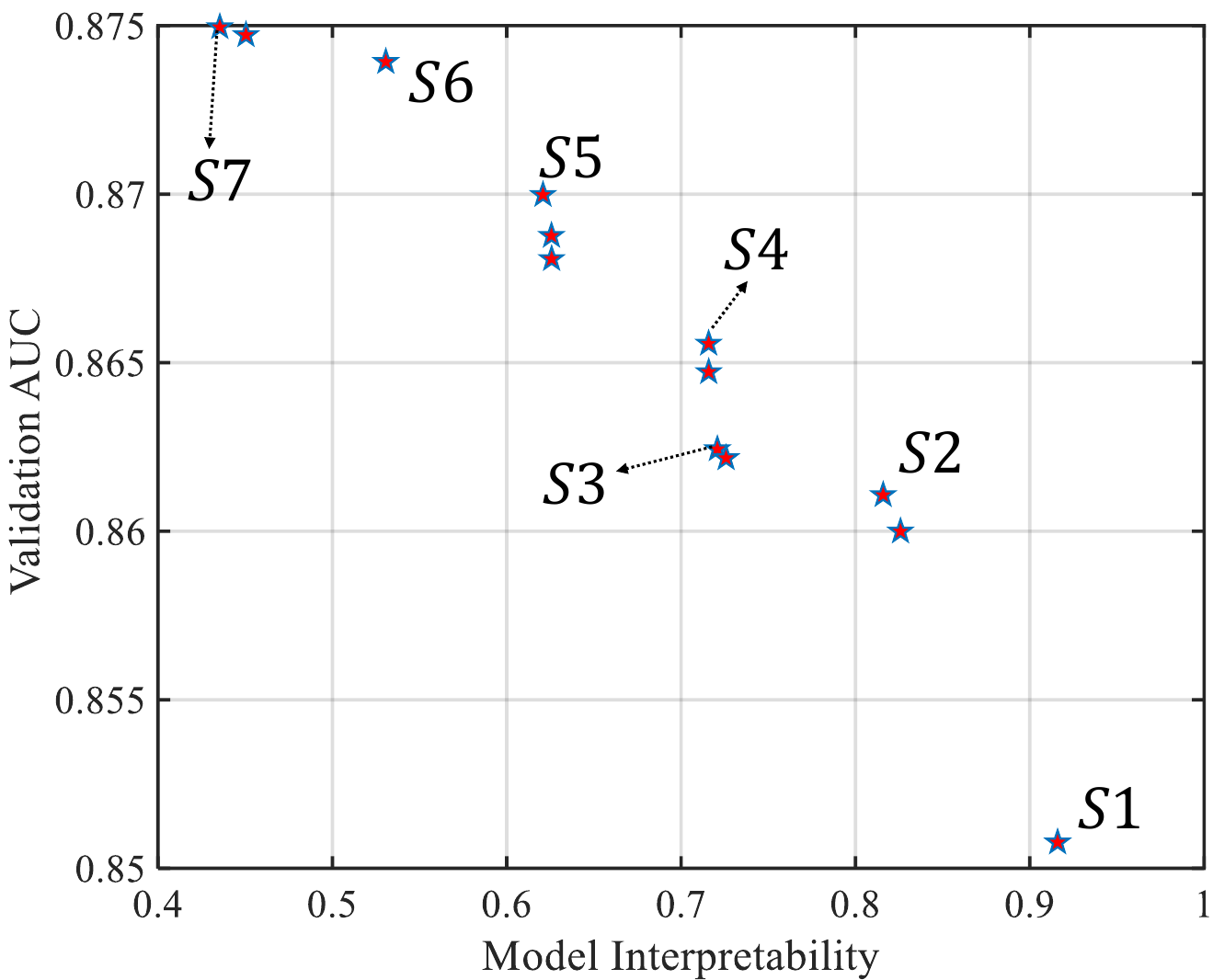}}
	\subfloat{\includegraphics[width=0.5670\linewidth]{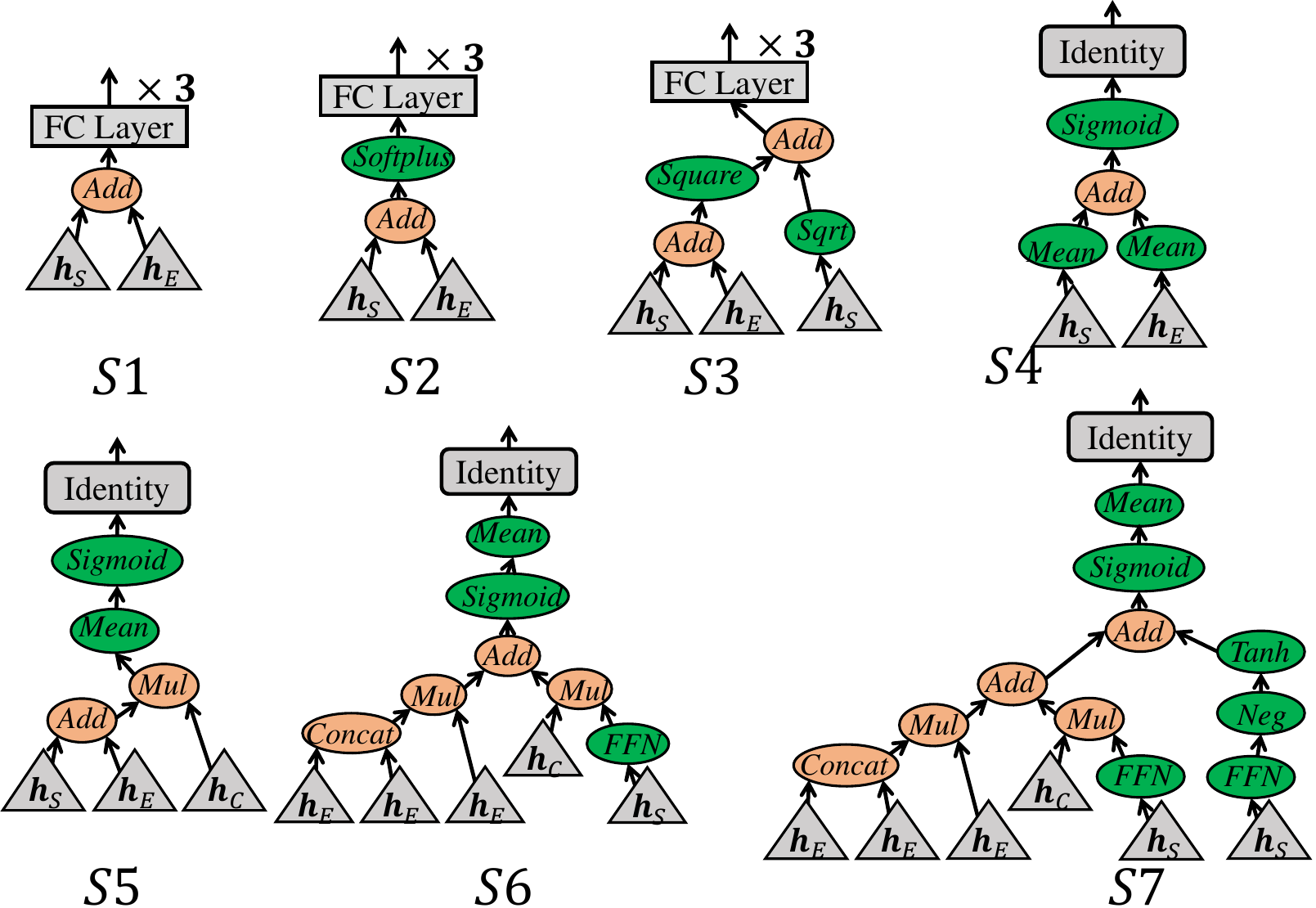}}	
	\caption{The non-dominated individuals found by the proposed EMO-NAS-CD on the SLP and the  visualization of architecture $S1$ to architecture $S7$.		% which  correspond to the architectures from EMO-NAS-CD-S1 to EMO-NAS-CD-S7, respectively.	
	}
	\label{fig:paretoslp}
\end{figure}

% 比性能
%

For a deep insight into  found architectures, we presented all non-dominated individuals found by the proposed approach on two datasets  in Fig.~\ref{fig:paretoASSIST} and Fig.~\ref{fig:paretoslp}, 
where the architectures corresponding to these individuals are further plotted in the right parts of  two figures.
%Architectures $A1$ to $A7$ represent architectures from EMO-NAS-CD-A1 to EMO-NAS-CD-A7, 
%while architectures $S1$ to $S7$ represent architectures from EMO-NAS-CD-S1 to EMO-NAS-CD-S7.
As can be observed, 
$A1$ or $S1$ is  the  shallowest architecture, which holds the highest model interpretability but worse prediction performance,
while $A7$ or $S7$ is the deepest  architecture, which holds the best performance but the worst interpretability among all selected architectures.

In addition, we can obtain some interesting and insightful observations  from these best-found architectures on two datasets.
%First, in ASSISTments dataset, architectures $A2$, $A3$, $A4$, and $A5$ are  similar, 
%and the difference between $A6$ and $A7$ is the \textit{Sqrt}  operation replaced by \textit{Abs}, 
%\XG{while  $A7$ has an extra \textit{Neg}  operation;}
%\XG{in the SLP dataset, architectures $S1$ to $S5$  have high similarities to each other,} and it seems that $S7$ is a variant of $S6$ with an extra branch. 
Firstly, from the comparisons of  $S1$ and $S2$, $A2$ and $A3$, as well as $A4$ and $A5$, we can find that adding a proper activation such as \textit{Sigmoid} and \textit{Softplus} can enhance the model performance 
without losing interpretability;
Secondly, in most  shallower architectures, the exercise-related input $\mathbf{h}_E$ tends to be directly combined with  the  student-related input $\mathbf{h}_S$ by some binary operators,
while in  most  deeper architectures,  $\mathbf{h}_E$ tends to be first combined with the knowledge concept-related input $\mathbf{h}_C$ and then combined with $\mathbf{h}_S$.
Finally,   all shallower architectures prefer FC layers  as their second parts to output the final prediction,
while for the deeper architectures with better performance,  
the \textit{Identity} operation seems to be a more effective second part.
These deeper  architectures  commonly obtain the final prediction with the assistance of the $Mean$ operator.
The above observations provide some valuable guidelines for manually designing novel CDMs.

\begin{table}[t]
	
	\centering
			\renewcommand{\arraystretch}{0.2}
		\caption{The performance comparison between the architectures found by EMO-NAS-CD on the ASSISTments and 
		the architectures found by EMO-NAS-CD on the SLP. The best results in each column are highlighted. Results are obtained on test datasets.}
	%\caption{The Performance Comparison between The Architectures Found by EMO-NAS-CD on The ASSISTments and 		The Architectures Found by EMO-NAS-CD on The SLP. The Best Results in Each Column are Highlighted.}
	\setlength{\tabcolsep}{0.5mm}{
	\begin{tabular}{c|ccc|ccc}
		\toprule
		\multicolumn{1}{c}{\textbf{Datasets}} & \multicolumn{3}{c}{\textbf{ASSISTments}} & \multicolumn{3}{c}{\textbf{SLP}} \\
		\midrule
		\multicolumn{1}{c}{\textbf{Methods}} & \textbf{ACC}$\uparrow$ & \textbf{RMSE}$\downarrow$ & \multicolumn{1}{c}{\textbf{AUC}}$\uparrow$ & \textbf{ACC}$\uparrow$ & \textbf{RMSE}$\downarrow$  & \textbf{AUC}$\uparrow$ \\
		\textbf{\textit{A2}} & 0.7452  & 0.4272  & 0.7726  & 0.7865  & 0.4030  & 0.8415  \\
	\textbf{\textit{A3}}  & 0.7353  & 0.4252  & 0.7766  & 0.7867  & 0.4001  & 0.8365  \\
		\textbf{\textit{A4}}  & 0.7857  & 0.4001  & 0.8210  & 0.7718  & 0.4105  & 0.8385  \\
	\textbf{\textit{A5}}  & 0.7810  & 0.4107  & 0.8409  & 0.7777  & 0.3924  & 0.8525  \\
	\textbf{\textit{A6}}  & 0.7884  & 0.4018  & 0.8474  & 0.7671  & 0.3966  & 0.8445  \\
		\textbf{\textit{A7}}  & \hl{0.7916 } & \hl{0.3993 } & \hl{0.8496 } & 0.7722  & 0.4007  & 0.8356  \\
		\midrule
		\textbf{\textit{S2}} & 0.7512  & 0.4788  & 0.7880  & 0.7915  & 0.3912  & 0.8560  \\
		\textbf{\textit{S3}} & 0.7435  & 0.4942  & 0.7786  & 0.7855  & 0.3912  & 0.8572  \\
		\textbf{\textit{S4}} & 0.7310  & 0.4218  & 0.7706  & 0.7964  & 0.3797  & 0.8635  \\
		\textbf{\textit{S5}} & 0.7355  & 0.4233  & 0.7736  & 0.7962  & 0.3794  & 0.8641  \\
		\textbf{\textit{S6}} & 0.7338  & 0.4281  & 0.7781  & 0.8001  & \hl{0.3773 } & 0.8669  \\
		\textbf{\textit{S7}} & 0.7321  & 0.4274  & 0.7769  & \hl{0.8023 } & 0.3781  & \hl{0.8685 } \\
		\bottomrule
\end{tabular}}
\\

	\label{tab:transfer}%
\end{table}%

\subsection{Architecture Transferring Validation}
As can be seen from Figs.~\ref{fig:paretoASSIST} and~\ref{fig:paretoslp}, 
the two  sets of selected best-architectures on two datasets are a bit different from each other. 
Only architecture $A1$ is same as architecture $S1$ and similar to $S2$ and $S4$,  and  architectures $A6$ and $A7$ are similar to architecture $S5$.

To further investigate the transferability and generalization of the found architectures,
Table~\ref{tab:transfer} presents the performance of architectures \textit{A2} to \textit{A7} and  
 architectures \textit{S2} to \textit{S7} on the two datasets under the splitting setting of 80\%/20\%, 
 where the results of \textit{A1} and \textit{S1} are not contained since they have  the same architecture.
As can be observed, 
the architectures found on the ASSISTments still hold  competitive performance on the SLP; 
similarly, the architectures found on the SLP also hold  comparable performance on the ASSISTments.
Note that architecture \textit{A5} and \textit{S2} have the best generalization to 	hold  the most promising performance on both  datasets.

\begin{figure}[t]
	\centering
	\includegraphics[width=0.6\linewidth]{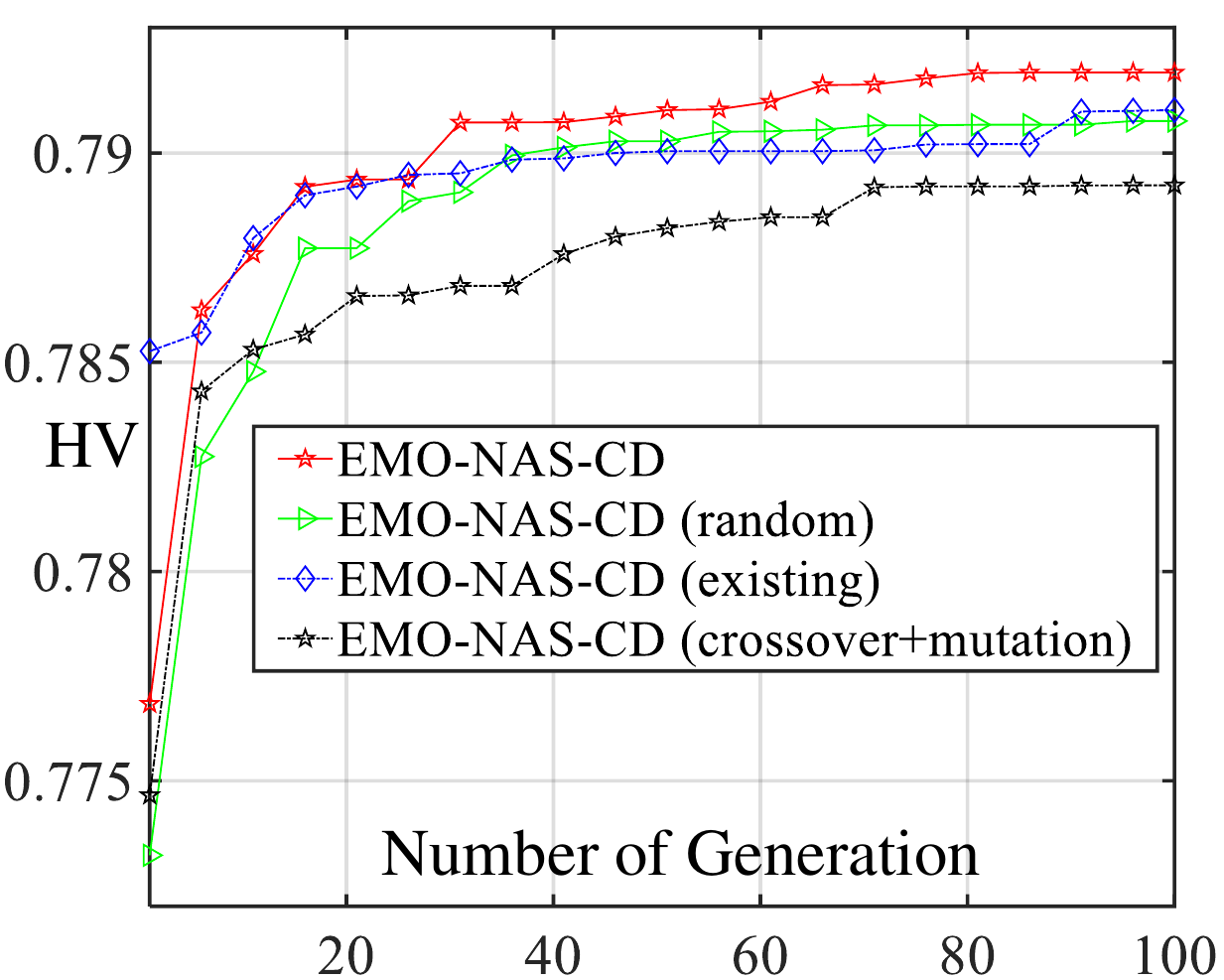}
	\caption{ Convergence profiles of HV obtained by EMO-NAS-CD and its variants on the SLP dataset.
		% where  one variant employs another type of genetic operator, and two variants are equipped with two different  initialization strategies respectively. 
	}
	\label{fig:hv}
\end{figure}

\begin{figure}[t]
	\centering
	\subfloat[EMO-NAS-CD with $f_2^{com}$.]{\includegraphics[width=0.515\linewidth]{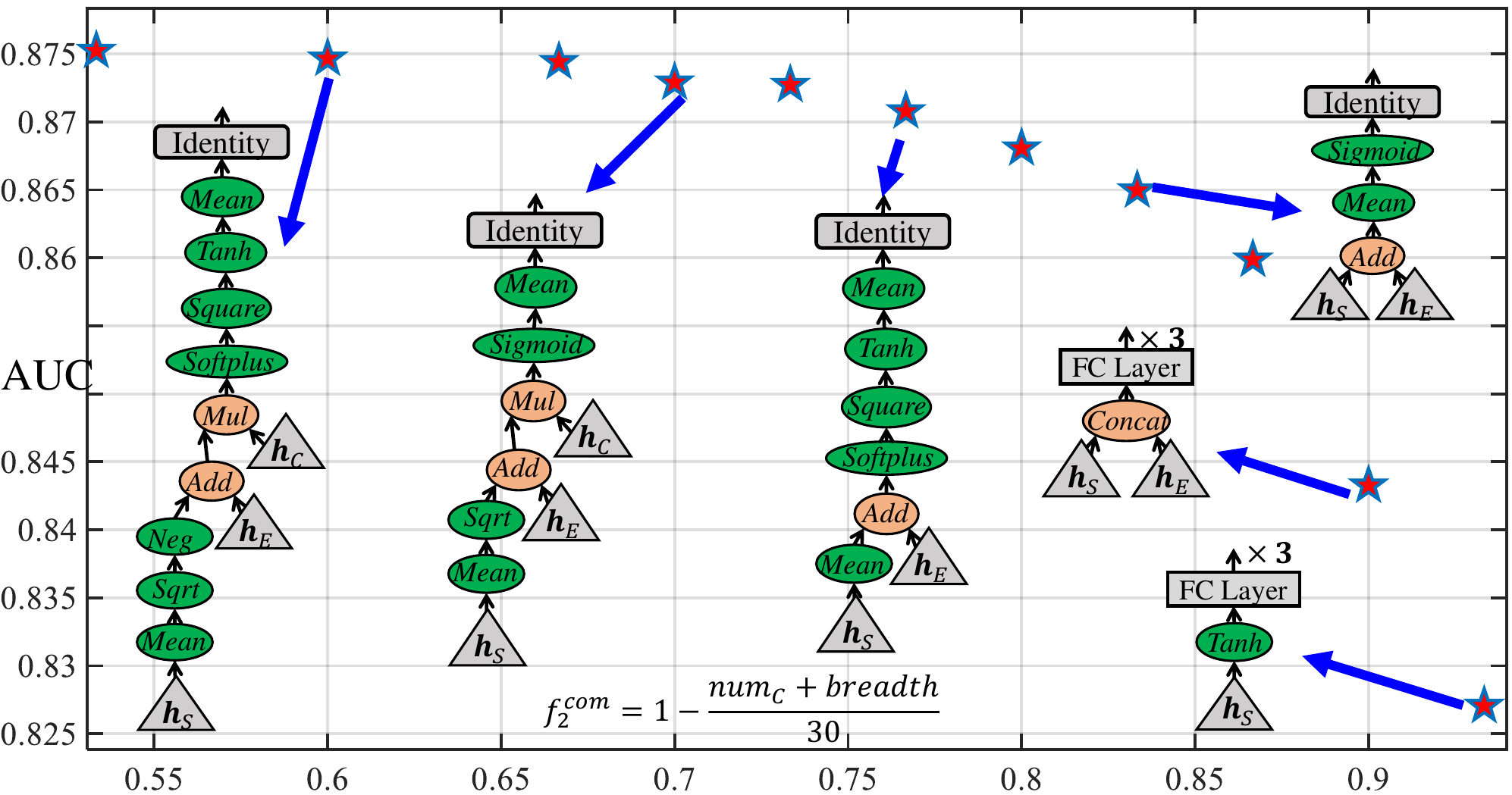}}
	\subfloat[EMO-NAS-CD with $f_2^{com\_dep}$.]{\includegraphics[width=0.485\linewidth]{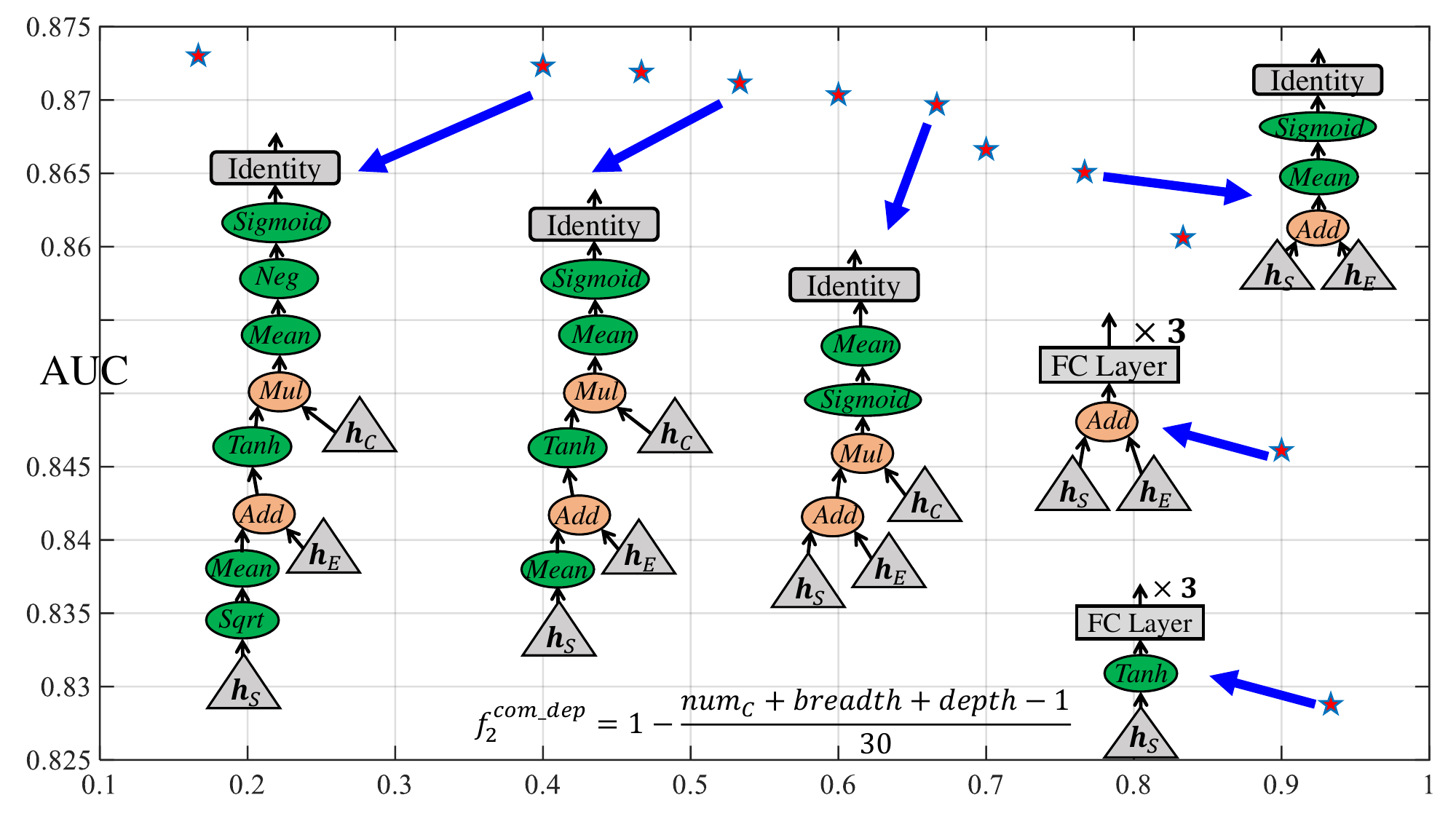}}
	\caption{The non-dominated individuals found by the proposed approach with $f_2^{com}$ and $f_2^{com\_dep}$, 
			and the architecture visualization of six representative  individuals. }
	\label{fig:f2slp}
\end{figure}

\begin{figure}[t]
	\centering
	\includegraphics[width=0.9\linewidth]{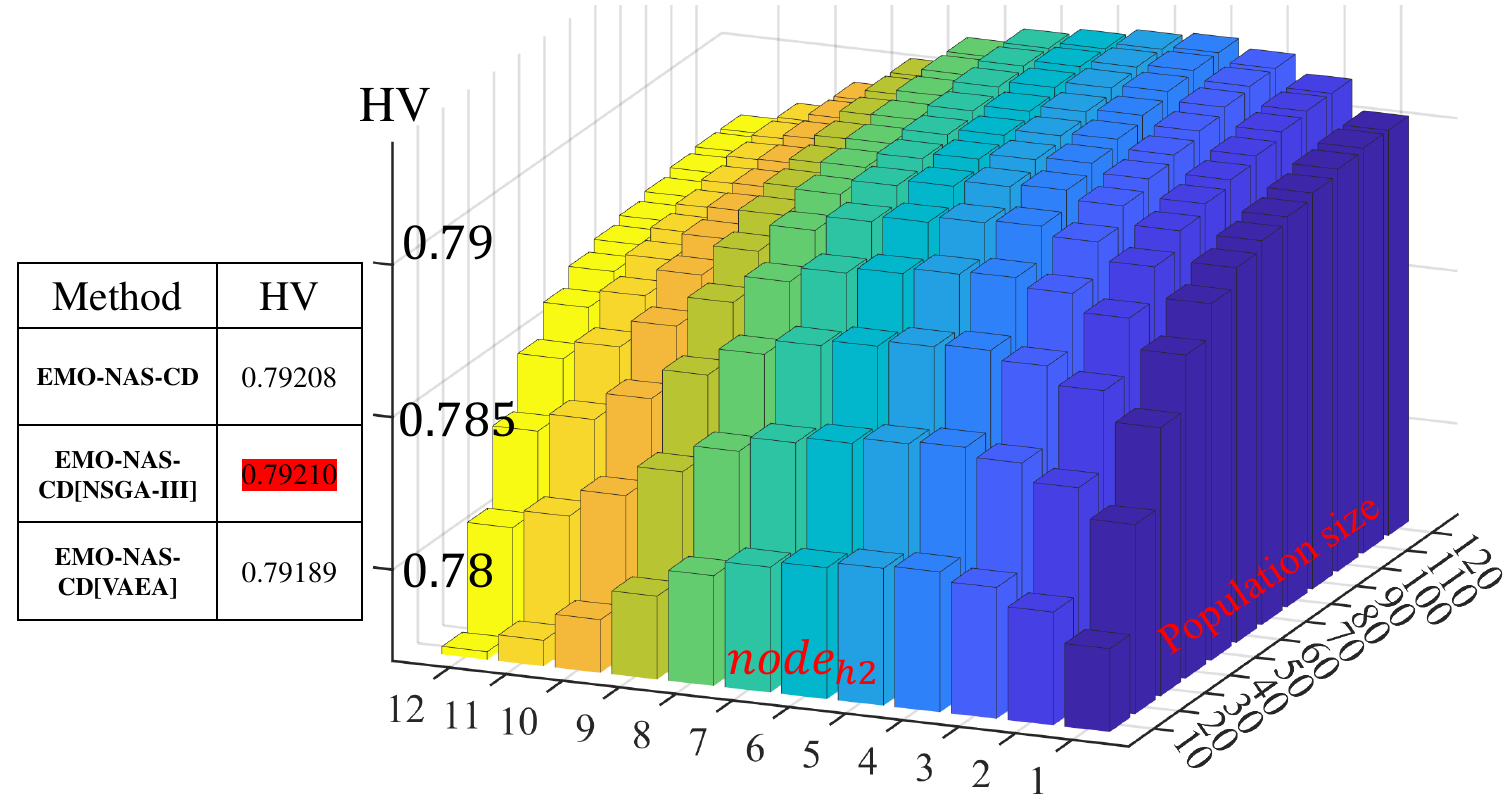}
	\caption{The influence analysis of hyperparameters ($Pop$ and $Node_{range}$)  on EMO-NAS-CD by the HV values.}
	\label{fig:hyperpara}
\end{figure}

\subsection{Ablation Study}
This section will validate the effectiveness of some devised strategies and analyze the parameter sensitivity.
In the following, only  the results on the SLP dataset is presented due to higher search cost on the ASSISTments.

To verify the effectiveness of the proposed initialization strategy, 
we equipped the proposed approach with other two  initialization strategies to form two variant approaches, \textit{EMO-NAS-CD (random) }and \textit{EMO-NAS-CD (existing)}.
The initial population of the former is randomly generated, and the latter initializes its population purely from existing CDMs.
Besides, we also established another variant called \textit{EMO-NAS-CD (crossover+mutation)}, which  generates offspring by first applying  the $Exchange$ operation and then randomly applying  one of other three operations.  
As a result, Fig.~\ref{fig:hv} presents the  convergence profiles of hypervolume (HV)~\cite{while2006faster} obtained by 
the proposed approach and its variants.
HV measures convergence and diversity of a population, and a large HV value indicates a good convergence and diversity.
 The comparison between EMO-NAS-CD and other two variants indicates that the suggested population initialization strategy can indeed speed up the convergence and lead to  better final convergence.
Besides, we can  observe that the proposed genetic operation is significantly better for the proposed approach than the compared genetic operation.
The reason behind this is that   successively employing two sub-genetic operations to generate offspring will cause  major modifications between the generated individuals and   their parent individuals, 
 which can promote the exploration of the algorithm  but hinder the exploitation of the algorithm  to some extent.
 To sum up, the effectiveness of the proposed  initialization and genetic operation can be demonstrated.

\iffalse
\begin{table}[t]
	\footnotesize
	\renewcommand{\arraystretch}{1}
	
	\centering
	\caption{  Final HV values of  the proposed EMO-NAS-CD and its variants on the SLP datasets. }
	\setlength{\tabcolsep}{0.0mm}{
		\begin{tabular}{cccc}
			\toprule
			\midrule
			\textbf{Method} & HV   & \textbf{Method}  & HV \\
			\midrule	
			\textbf{EMO-NAS-CD} &  \textbf{ 0.79208}    &-   &  -\\
			\midrule	
			\midrule	
			
			\textbf{EMO-NAS-CD[NSGA-III]} &   \hl{0.79210}   & \textbf{EMO-NAS-CD[VAEA]}      & 0.79189 \\
			\midrule
			\midrule
			\textbf{EMO-NAS-CD(Pop=20)} &  0.78513     & \textbf{EMO-NAS-CD(Pop=50)}     & 0.79058 \\
			\textbf{EMO-NAS-CD(Pop=80)}  &   0.79172    &  \textbf{EMO-NAS-CD(Pop=120)}    & \hl{0.79215}\\
			\midrule
			\midrule		
			\textbf{EMO-NAS-CD(1\_2)} &   0.79021    & \textbf{EMO-NAS-CD(2\_6)}     &  0.79163\\
			\textbf{EMO-NAS-CD(2\_8)}  &  0.79103     &  \textbf{EMO-NAS-CD(2\_10)}  & 0.78923 \\
			\bottomrule

		\end{tabular}%
	}
	\label{tab:ablation}%
\end{table}%
\fi

To validate the effectiveness of  objective $f_2(\mathcal{A})$ of (\ref{eqa:f2}) in assisting the proposed approach to search interpretable CDMs,
Fig.~\ref{fig:f2slp} exhibits the non-dominated individuals found by two variants of the proposed approach and plots 
their six representative architectures for observation.
Here, the first variant takes the model complexity as the  second objective: $f_2^{com} = 1-\frac{numc+breadth}{30}$, 
and the second variant computes the model complexity  as $f_2^{com\_dep} = 1-\frac{numc+breadth+depth-1}{30}$.
$f_2^{com}$ is measured by the sum of the numbers of computation nodes and leaf nodes, 
 and  $f_2^{com\_dep}$ additionally considers  the influence of the tree depth (30 is a parameter used for normalizing the objective value).
As can be seen from Fig.~\ref{fig:f2slp}(a),   compared to the  architectures in the right part, 
the architectures in the left part have  better performance but at the expense of a much larger increase of depth.
  Besides,   the architectures (located in the upper left area) are much deeper  compared to  the architectures with similar performance in Fig.~\ref{fig:paretoslp}.
The reason behind this  is that the objective of model complexity prefers adding a  unary operator node, whereas adding a binary operator node would introduce an extra leaf node, leading to a worse objective value.
 The same observation and conclusion can be drawn  from Fig.~\ref{fig:f2slp}(b), 
where the found architectures are still very deep. 
This is because $f_2^{com\_dep}$ is  basically the same as $f_2^{com}$ 
yet implicitly assigns a smaller penalty to binary operator nodes, 
where the assigned penalty is still larger than the penalty assigned to unary operator nodes  by $f_2^{com\_dep}$.
Finally, the effectiveness of the devised    model interpretability objective can be validated.

To analyze the sensitivity of   the proposed approach to the framework of MOEAs and  the hyperparameters $Pop$ and $Node_{range}$, 
	Fig.~\ref{fig:hyperpara} compares HV values on the SLP  obtained by EMO-NAS-CD  under different hyperparameter combinations of  $Pop$ and $Node_{range}$.
	According to Taguchi method~\cite{sadeghi2015two}, $Pop$ is  set from 10 to 120 with  step equal to 10, while $node_{h2}$ in $Node_{range}$ is set from 1 to 12 with  step equal to 1  and	 $node_{h1}$ is fixed to 2.
	The original EMO-NAS-CD is under  NSGA-II, 	but EMO-NAS-CD[NSGA-III] and EMO-NAS-CD[VAEA] are  EMO-NAS-CD under NSGA-III~\cite{deb2013evolutionary} and VAEA~\cite{xiang2016vector}, respectively.
	As can be seen from Fig.~\ref{fig:hyperpara}, firstly,  the proposed EMO-NAS-CD is robust to the framework of MOEAs;	  
	secondly, the proposed EMO-NAS-CD can obtain  relatively good performance when the population size is greater than 80, 
	and it is not necessary to set $Pop$ to 120 for a slightly higher HV value at the expense of an extra 0.2 times of cost;
	thirdly, the setting of  $node_{h2}$ has a big influence on the result of the  EMO-NAS-CD, 
	and  	 EMO-NAS-CD can obtain relatively good performance when $node_{h2}$  lies  from 3 to 5.
	Therefore, current hyperparameter settings for EMO-NAS-CD are  good enough to some extent.

\subsection{Discussion}
This section will discuss  three  guidelines for researchers in various domains after the experiments. 

The first guideline is for researchers in NAS. 
To design a task-specific NAS approach, researchers should make the best of their domain knowledge to create a search space. By doing so, the search space can include existing models for the target task and many other potential models.  In addition, the search strategy should also be based on the search space's characteristics and the target task's domain knowledge.
The second guideline is for researchers in CD, inspiring them on how to design effective CDMs, where the detailed guideline can be found in the last paragraph of Section~\ref{sec:Eff}. 
The third guideline is for researchers interested in NAS and intelligent education. 
Considering the success made by our approach, it is promising for other tasks in intelligent education to employ the NAS technique to design effective neural architectures.
Besides, researchers can borrow experiences from this paper to design the objectives of model interpretability,  generalization, and robustness,  formulate their multiple objectives as a MOP, and then employ a suitable MOEA to solve the MOP.

\iffalse

\begin{figure}[t]
	\centering
	\includegraphics[width=1\linewidth]{f2_SLP.pdf}
	\caption{\XG{The non-dominated individuals searched by the proposed approach with $f_2^{com}$ and the visualization of six individuals' architectures. }}
	\label{fig:f2slp}
\end{figure}

\begin{figure}[t]
	\centering
	\includegraphics[width=1\linewidth]{pareto_SLP_f2_c_depth.pdf}
	\caption{\XG{The non-dominated individuals searched by the proposed approach with $f_2^{com_dep}$ and the visualization of six individuals' architectures. }}
	\label{fig:f2slp_c}
\end{figure}
\begin{figure*}[t]
	\centering
	\subfloat[$f_2^{com}$.]{\includegraphics[width=0.5\linewidth]{f2_SLP.pdf}}
	\subfloat[$f_2^{com\_dep}$.]{\includegraphics[width=0.5\linewidth]{pareto_SLP_f2_c_depth.pdf}}
	\caption{\XG{The non-dominated individuals searched by the proposed approach with $f_2^{com}$ and the visualization of six individuals' architectures. }}
	\label{fig:f2slp}
\end{figure*}

\fi

\section{Conclusion and Future Work}
In this paper, we proposed to design  novel CDMs by leveraging evolutionary multi-objective NAS.
Specifically, we first proposed an expressive   search space for CD, 
which  contains a large number of potential architectures and existing architectures.
Then, we proposed an effective MOGP to search high-performance architectures with high interpretability in the search space by optimizing the MOP having two objectives.
To avoid some optimization difficulties,  each architecture is first transformed into its corresponding tree architecture and then encoded by tree-based representation for easy optimization.
Besides, in the proposed MOGP, an effective genetic operation is designed for offspring generation, 
and a population initialization strategy is devised to accelerate the convergence.
Experimental results demonstrate the superiority of the architectures found by the proposed approach to existing CDMs in terms of performance and model interpretability.

This work has shown the promising prospect of leveraging NAS for CD, but there still exist some threats to the validity of the proposed approach, including internal, external, and construct threats.
Firstly, the  devised model interpretability objective is the primary  internal threat. 
As seen from Fig.~\ref{fig:objectives}, Fig.~\ref{fig:paretoASSIST}, and Fig.~\ref{fig:paretoslp}, 
the proposed approach will find relatively different architectures when different model interpretability objectives are adopted. Besides, the proposed model interpretability objective is empirically designed based on some experiences from decision trees, which may limit the emergence of novel architectures as well as the application of found architectures in the real world due to a little unreasonable definition of model interpretability. 
Therefore,  we would like to design more reasonable model interpretability objectives in the future.
Secondly, the dataset utilized in the proposed approach is the main external threat.
We can find that the found architectures on different datasets are quite different, which indicates that the architectures found by the proposed approach on a single dataset are not general for the cognitive diagnosis task.
Besides, the size of the utilized dataset affects the search efficiency of the proposed approach, 
which leads to an extremely high computation cost when a large-scale dataset is met, e.g., the search cost on ASSISTments is about 15 GPU days.
Therefore, in the future, we would like to  design generalized CDMs and explore  surrogate models~\cite{sun2019surrogate,Chen2022Neural} to reduce the search cost.
Finally,  the proposed search space is the main construct threat since it is designed based on the summary of existing architectures and forces all architectures to be single-root trees. Despite high effectiveness, the current search space may limit the emergence of more potential architectures since CDMs should not always be single-root trees. Therefore, it is interesting to devise other types of search space to contain more effective CDMs.

\iffalse
Firstly, the proposed approach suffers from high computation cost, e.g., the search cost on ASSISTments is about 15 GPU days, and it will become severer  when  searching  architectures for a  large-scale dataset, 
where the training time of each architecture is much more expensive.
Therefore,  in the future, we would like to explore  surrogate models~\cite{sun2019surrogate,liu2021surrogate} to reduce the search cost.
Moreover, our proposed  search space is designed based on the summary of existing architectures and forces all architectures to be single-root trees,
which is effective to some extent but may ignore many potential architectures.
Therefore,  it is interesting to devise a more effective search space  and encoding strategy  for CD.
\fi

%Yuanchao Liu received the master degree in control theory and control engineering
%from Northeastern University, Shenyang, China, in 2019. He is
%currently pursuing the Ph.D. degree of control science and
%engineering in Northeastern University, Shenyang, China.

%His current research interests include multiobjective optimization,
%robust optimization, dynamic optimization and data-driven
%optimization.

\bibliographystyle{IEEEtran}
\bibliography{NASCD}
\end{document}